\documentclass[10pt,twocolumn,letterpaper]{article}

\usepackage[pagenumbers]{cvpr} % To force page numbers, e.g. for an arXiv version

\usepackage{graphicx}
\usepackage{amsmath}
\usepackage{amssymb}
\usepackage{booktabs}
\usepackage{multirow}
\makeatletter
\@namedef{ver@everyshi.sty}{}
\makeatother
\usepackage{tikz}
\usepackage[outline]{contour}
\usepackage{placeins}
\usepackage{tikz}
\usepackage{pgfplots}
\usepackage{microtype}
\usepackage{subcaption}
\usepackage{siunitx}

\newcommand{\proj}{\text{proj}}

\DeclareMathOperator*{\upsample}{upsample}
\DeclareMathOperator*{\dosample}{downsample}

\usepackage[pagebackref,breaklinks,colorlinks]{hyperref}

\usepackage[capitalize]{cleveref}
\crefname{section}{Sec.}{Secs.}
\Crefname{section}{Section}{Sections}
\Crefname{table}{Table}{Tables}
\crefname{table}{Tab.}{Tabs.}

\def\cvprPaperID{17} % *** Enter the EarthVision Paper ID here
\def\confName{EarthVision}
\def\confYear{2022}

\begin{document}

\definecolor{cadetgrey}{rgb}{0.57, 0.64, 0.69}
\definecolor{gold(metallic)}{rgb}{0.83, 0.69, 0.22}
\definecolor{carrotorange}{rgb}{0.93, 0.57, 0.13}
\definecolor{brown}{rgb}{0.59, 0.29, 0.0}
\definecolor{cadetblue}{rgb}{0.37, 0.62, 0.63}
\definecolor{darkolivegreen}{rgb}{0.33, 0.42, 0.18}
\definecolor{applegreen}{rgb}{0.55, 0.71, 0.0}
\definecolor{bananayellow}{rgb}{1.0, 0.88, 0.21}
\definecolor{arylideyellow}{rgb}{0.91, 0.84, 0.42}
\definecolor{blue(ncs)}{rgb}{0.0, 0.53, 0.74}
\definecolor{indigo}{rgb}{0.29, 0, .51}
\definecolor{bleudefrance}{rgb}{0.19, 0.55, 0.91}

\title{Multi-Layer Modeling of Dense Vegetation from Aerial LiDAR Scans}

\author{
Ekaterina Kalinicheva\textsuperscript{1, 2}
\and
Loic Landrieu\textsuperscript{1}
\and
Clément Mallet\textsuperscript{1}
\and
Nesrine Chehata\textsuperscript{1,3}
\and
{\textsuperscript{1}Université Gustave Eiffel, ENSG, IGN, LASTIG, F-77454 Marne-la-Vallee, France}
\\
{\textsuperscript{2}INRAE, UMR 1202 BIOGECO, Université de Bordeaux, France}
\\
{\textsuperscript{3}Bordeaux INP, Université Bordeaux Montaigne, France}\\
{\tt\small ekaterina.kalinicheva@ign.fr}
}

% \author{Ekaterina Kalinicheva\\
% IGN\\
% {\tt\small ekaterina.kalinicheva@ign.fr}
% \and
% Loic Landrieu\\
% {\tt\small 	loic.landrieu@ign.fr}
% \and
% Clément Mallet\\
% {\tt\small 	clement.mallet@ign.fr}
% }
\maketitle

%%%%%%%%% ABSTRACT
\begin{abstract}
    The analysis of the multi-layer structure of wild forests is an important challenge of automated large-scale forestry.
    While modern aerial LiDARs offer geometric information across all vegetation layers, most datasets and methods focus only on the segmentation and reconstruction of the top of canopy.
    We release WildForest3D, which consists of $29$ study plots and over $2000$ individual trees across $47\,000$m$^2$ with dense 3D annotation, along with occupancy and height maps for 3 vegetation layers: ground vegetation, understory, and overstory.
    We propose a 3D deep network architecture predicting for the first time both 3D point-wise labels and high-resolution layer occupancy rasters simultaneously. This allows us to produce a precise estimation of the thickness of each vegetation layer as well as the  corresponding watertight meshes, therefore meeting most forestry purposes. {Both the dataset and the model are released in open access: \url{https://github.com/ekalinicheva/multi_layer_vegetation}}.
    % We propose a new deep learning-based algorithm for prediction the occupancy and height of different vegetation strata from airborne 3D LiDAR point clouds. Using partially annotated data, our model
    %firstly classify 3D points into different vegetation classes, and then projects them to the corresponding 2D occupancy maps of different vegetation layers - ground vegetation, understory and overstory - along with the height of vegetation coverage.
    %We provide the first 3D point cloud database containing around 2000 annotated trees and bushes spread over 29 study plots. In our work, we that by generating the supplementary 2D training data it becomes possible to identify classes that were not annotated, and, hence, drastically increase the quality of the prediction. To our knowledge, our method is unique in the forestry domain as it produces the high-precision fine resolution occupancy maps of different strata along with their respectful height, so that the 3D vegetation data can be presented in a form of mesh for each vegetation layer.
\end{abstract}

%%%%%%%%% BODY TEXT
%-------------------------------------------------------------------------
\section{Introduction}
\label{sec:intro}
The retrieval and analysis of the multiple layers of dense vegetation is an essential step for many forestry and ecology applications, such as forest management~\cite{jantunen2001effects}, biodiversity and habitat analysis~ \cite{gaujour2012factors,mapping_snags}, biomass estimation \cite{ferraz2016airborne,lu2006potential}, or forest fire modeling~\cite{mckenzie201115, maclean1996forest, sandberg2001characterizing}.
Thanks to the constant advances in remote sensing technology, we can now gather large amounts of precise geometric and radiometric data on vast forested environments.
However, standard satellite or aerial images are only suitable for the analysis of the top canopy layer \cite{TUANMU20101833}.
Conversely, terrestrial laser scanning (TLS) captures detailed information on the bottom layers with slightly decreasing point density toward the upper vegetation layers. 
Unfortunately, TLS data acquisition over large areas ($>1\:$ha) is unpractical due to the high amount of resource involved~\cite{CALDERS2020112102}.
In contrast, Airborne Laser Scanners (ALS), and especially Unmanned Aerial Vehicles equipped with Laser Scanners (UAV-LS), can capture 3D point clouds over larger areas with sufficient point density~\cite{CALDERS2020112102, brede2017comparing, liang2019forest}.

Occlusion of the overstory tree cover leads to a limited 3D sampling of the bottom vegetation layers: their analysis from an airborne LiDAR perspective remains challenging, and most vegetation analysis algorithms only explicitly model the visible vegetation layers. Moreover, given the difficulty of annotating the lower strata of dense forests, very few annotated and public datasets exist for the training and evaluation of approaches modeling the multi-layer structure of dense forests. Large-scale assessment is also compromised.

\begin{figure}
    \centering
    \resizebox{\linewidth}{!}{
    \begin{tabular}{lllllll}
    \multicolumn{7}{c}{
    \includegraphics[width=1.2\linewidth]{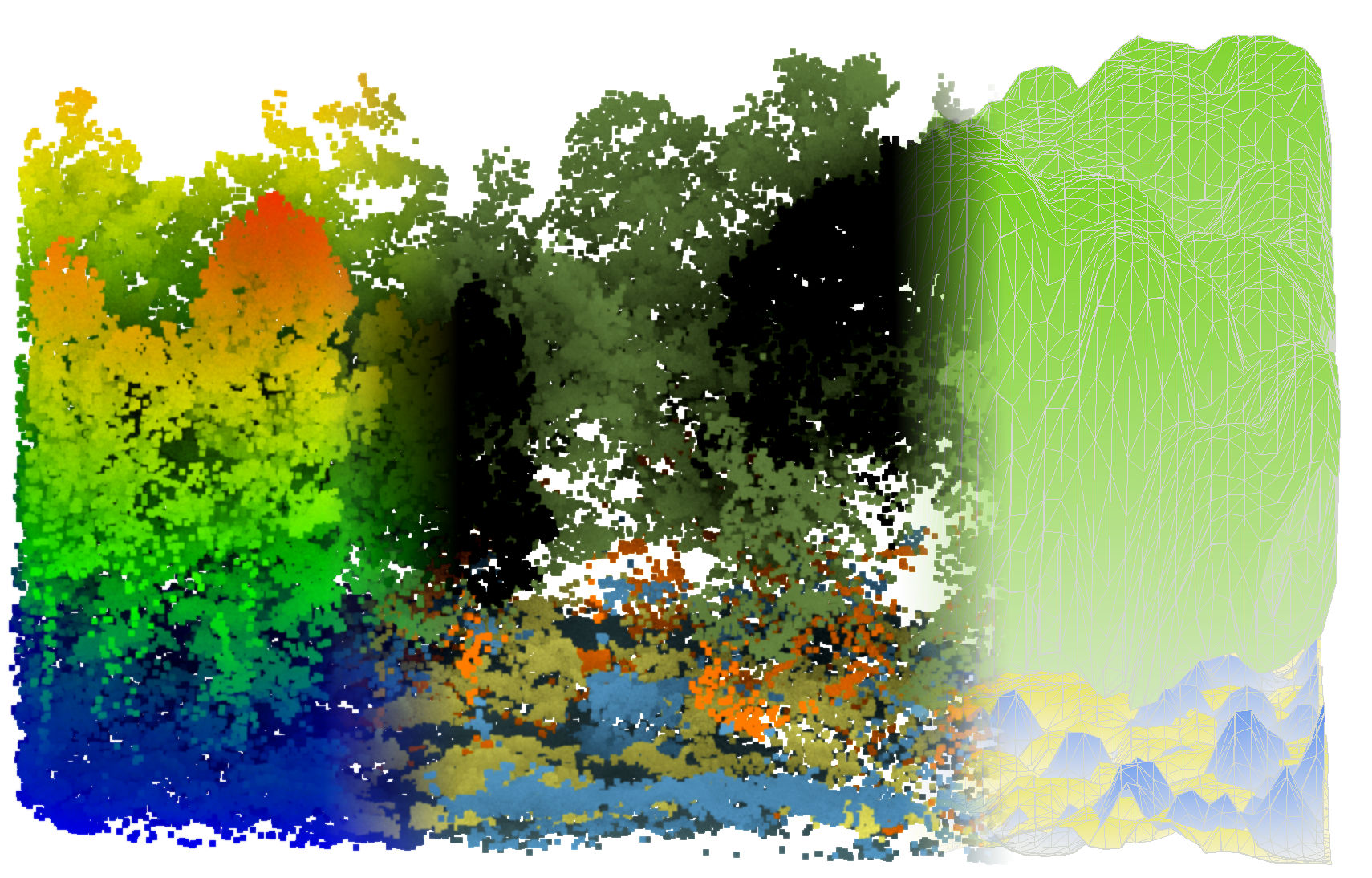}
    }\\
    \multicolumn{4}{c}{point labels} & &
    \multicolumn{2}{c}{vegetation layers} \\
    \cline{1-4} \cline{6-7}
    ~\vspace{-3mm}\\
    
    \tikz \fill[fill=darkolivegreen, scale=0.3, draw=black] (0,0) rectangle (1,1); 
    & deciduous &
    \tikz \fill[fill=black, scale=0.3, draw=black] (0,0) rectangle (1,1); 
    & coniferous &\;&
    \tikz \fill[fill=applegreen, scale=0.3, draw=black] (0,0) rectangle (1,1); 
    & overstory \\
    
    \tikz \fill[fill=carrotorange, scale=0.3, draw=black] (0,0) rectangle (1,1); 
    & stem &
    \tikz \fill[fill=blue(ncs), scale=0.3, draw=black] (0,0) rectangle (1,1); 
    & understory \; \; \; &&
    \tikz \fill[fill=bleudefrance, scale=0.3, draw=black] (0,0) rectangle (1,1); 
   & understory \\ 
   
   \tikz \fill[fill=arylideyellow, scale=0.3, draw=black] (0,0) rectangle (1,1); 
   & ground vegetation &
   \tikz \fill[fill=cadetgrey, scale=0.3, draw=black] (0,0) rectangle (1,1); 
   & ground &&
   \tikz \fill[fill=yellow, scale=0.3, draw=black] (0,0) rectangle (1,1); 
   & ground veget.
    \end{tabular}
    }
    \caption{{\bf Overview.} We propose WildForest3D, a novel dataset of annotated UAV-LS point clouds of dense forest (left), and a new model for the multi-layer analysis of vegetation. Our network performs 3D semantic segmentation (middle), and produces height maps and watertight meshes for three vegetation layers (right).}
    \label{fig:teaser}
\end{figure}

In this paper, we propose a first-of-its-kind dataset of $29$ plots of dense forest, $7$ million 3D points and $2.1\:$million individual labels. This corresponds to over $2000$ tree instances which were individually located and classified by \textit{in situ} observations from forestry experts. The labels indicate the nature of the vegetation structure containing the points: deciduous canopy, coniferous canopy, understory, stems, and ground. We use these annotations to produce high resolution maps of the occupancy and thickness of three vegetation layers: ground vegetation, understory, and overstory.

We also introduce a deep learning-based model for the automated analysis of multi-layer vegetation from aerial laser scans, see Figure~\ref{fig:teaser}. Our network operates directly on the 3D points to perform semantic segmentation of the point clouds and generate layer occupancy rasters that can be supervised end-to-end. We can then produce height maps for all layers, which we transform into watertight meshes. These outputs are useful for downstream applications such as biomass, carbon stock, fire fuel estimation \cite{ferraz20123,GALE2021112282}, soil illumination \cite{hardtle2003effects}, or vegetation parameters extraction for the forest inventory~\cite{rs13245113}.
The contributions of this paper are as follows:
\begin{itemize}
    \item We present WildForest3D, the first 3D forest dataset with dense annotations, with  over $2000$ trees.\vspace{-0.15cm}
    \item We propose a deep learning-based method to model the multi-layer structure of dense forest from UAV-LS 3D point. %\EKAT{Our method can be used to generate high resolution height maps of three different vegetation layers.}
    Our method can be used to generate high resolution meshes of three different vegetation layers.
    \vspace{-0.15cm}
\end{itemize}
%\EKAT{Both dataset and model are released in open access at }.
%Our dataset and implementation will be released in open-access upon publication.
%.........................
\section{Related Work}
%.........................
The machine learning paradigm is still relatively new for forest data analysis. Most algorithms are based on the regression between point cloud features and \textit{in situ} metrics, which typically do not generalize well outside of a given study area. 
We first present an overview of the existing open datasets, then the traditional approaches for structured vegetation analysis, and finally the recent deep learning advances. 

\paragraph{\bf Existing Forestry Datasets.}
Detailed annotation of ALS forestry data is a labor-intensive process, requiring expert knowledge and often \textit{in situ} observations.
For this reason, ALS annotations are often aggregated at the plot-level \cite{ekaterina_kalinicheva_2021_5555758}, but restricted for forest inventory purposes. They have never been exploited for fine-grained multi-layer semantic segmentation. Numerous countries now provide publicly available low resolution ALS scans at large scale \cite{denmark,finland,sweden} but annotations are still lacking. This data can be combined with the open forest inventory EFISCEN \cite{EFISCEN}, containing annotations aggregated from plot to regions for 32 countries.

Several 3D LiDAR-based benchmarks have been released in the two last decades but they were focused on the specific task of individual tree crown delineation, i.e., a segmentation task over the canopy layer \cite{Wang_ITC}. Weinstein \etal \cite{ben_weinstein_2022_5914554} provide a first-of-its-kind dataset for canopy crown detection and delineation containing {co-registered airborne LIDAR point clouds, and RGB and hyperspectral images}. In total, $30\,975$ tree crowns are delimited with 2D bounding boxes.
Our proposed WildForest3D contains richer annotations, at the level of the individual 3D points, which required an extensive campaign of \textit{in situ} observations.

\paragraph{\bf Vegetation Structure Analysis from ALS Data.}
Automated analysis of vegetation structure can be divided into tree-based \cite{10.1093/forestry/cpr051, HAMRAZ2016532, VEGA201498, REITBERGER2009561} and area-based \cite{Latifi2016EstimatingOA, Generalizing_predictive_models_of_forest_inventory} methods. The former start by segmenting individual trees and aggregating their parameters over an area. The latter derive vegetation statistics from a study area and directly regress target variables. 
Both methods tend to overlook small elements of bottom layers due to the lower ALS point density {in bottom layers}, focusing on the analysis of higher vegetation, e.g., dominant trees. 

Despite their limited financial value \cite{Forest_understory_trees}, the lower vegetation layers are important for many ecological applications, such as the study of forest habitats  \cite{gaujour2012factors} or fire risk management \cite{mckenzie201115}. However, detecting these layers is comparatively more challenging than overstory tree canopy. Hamraz \etal \cite{Forest_understory_trees} report that while $90\%$ of overstory trees are correctly segmented by {most methods,}
%their method,
lower vegetation detection barely reaches $60$\% accuracy.
The lower vegetation analysis is often performed at an aggregated level over large areas. It consists in vegetation statistics at low-resolution level around $100-400m^2$ \cite{10.1371/journal.pone.0220096, mapping_snags, jarron2020detection}.
Ferraz \etal \cite{multilayer_crown_maps} model canopy cover of different vegetation layers using kernel-based estimators coupled with individual tree crown detection. Their approach proves particularly relevant for modeling the overstory, but is affected by the low point density of lower vegetation. 

% %Both methods tend to overlook the small vegetation of lower layers and struggle in high density wild forest with intersecting canopies. 

% As a consequence, most works focus on the overstory layer \cite{}, or only provide statistics on lower layer aggregated over large areas \cite{10.1371/journal.pone.0220096} or with low resolution, \eg $20$m for \cite{mapping_snags}.   

% Ferraz \etal \cite{multilayer_crown_maps} model crown cover for canopy density modelling using kernel-based estimators. Their approach proves particularly relevant for modeling the overstory, but is affected by the low density of lower vegetation. 
\begin{figure}[t]
  \centering   
%   \fbox{\rule{0pt}{2in} \rule{0.9\linewidth}{0pt}}
   \includegraphics[width=1\linewidth]{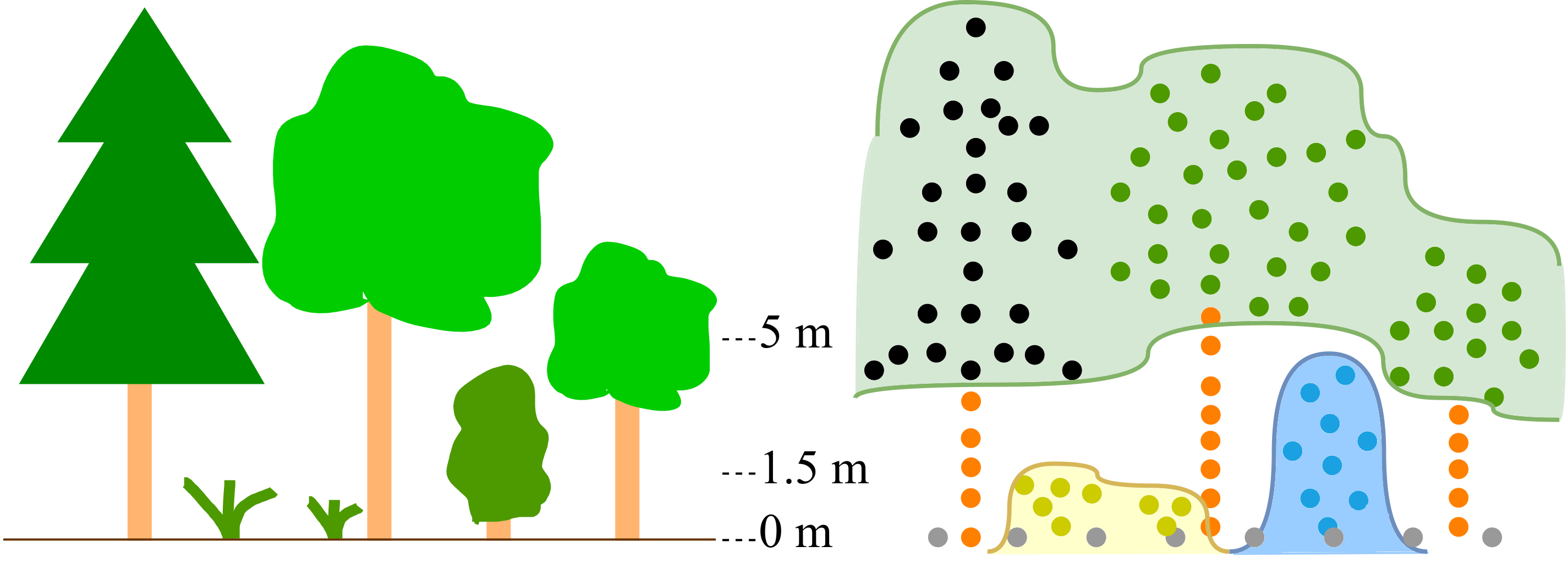}
  \caption{\textbf{Multi-Layer Structure.}
  We distinguish three vegetation layers: \textcolor{yellow!70!black}{ground vegetation}, \textcolor{bleudefrance}{understory}, and \textcolor{applegreen}{overstory}. Note that the layers are determined by the height of the vegetation instances they belong to, and not the point height: the bottom limit of the overstory layer may be lower than the top limit of the understory.}
   \label{fig:stratum_example}
\end{figure}

\begin{figure*}[t]
\centering
\begin{tabular}{@{}ccccc}
\begin{subfigure}{0.23\textwidth}
    \includegraphics[width=\linewidth,height=.13\textheight]{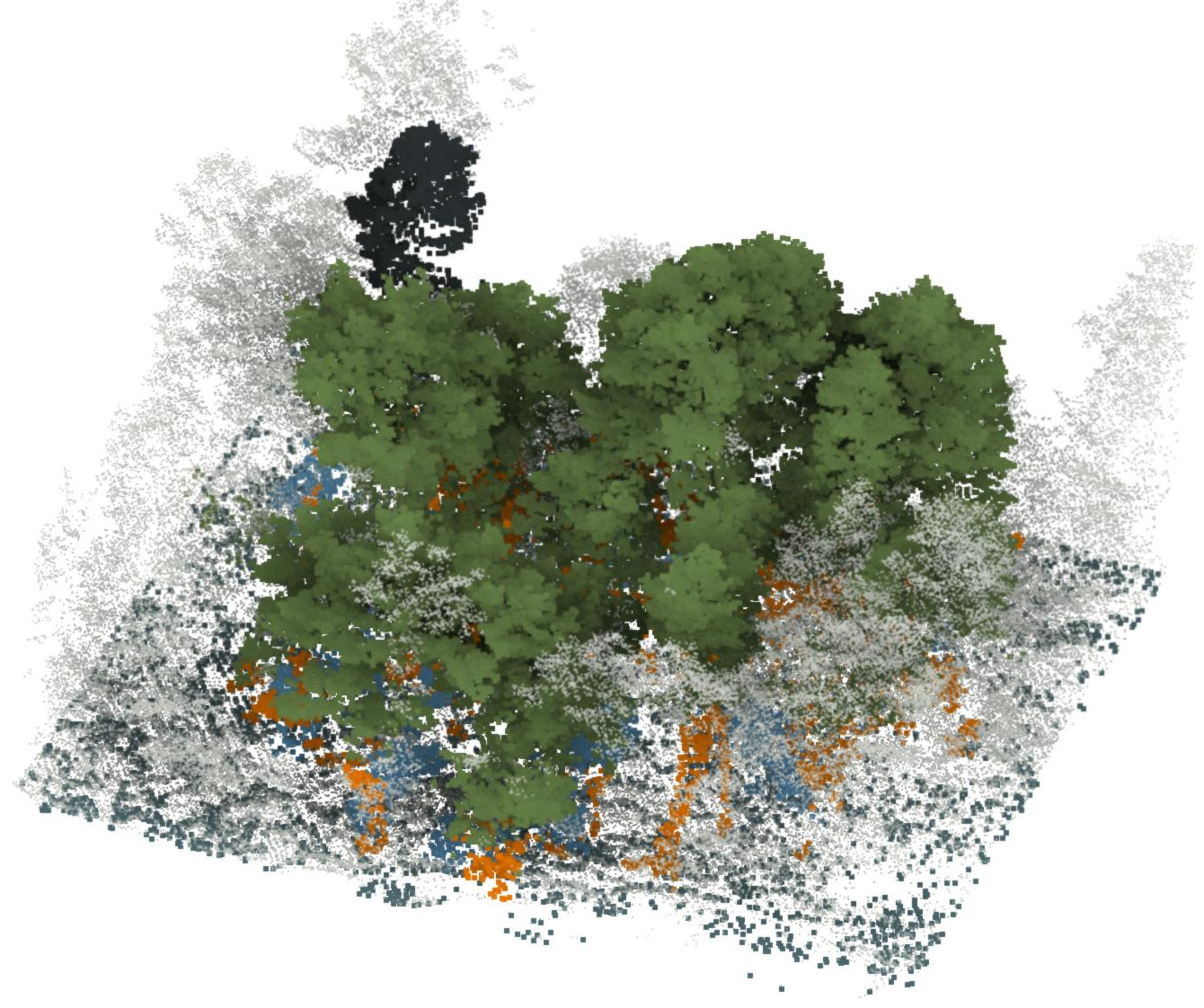}
    \caption{Annotated 3D Point Cloud.}
    \label{fig:3D_annot}
\end{subfigure}
&
\begin{subfigure}{0.17\textwidth}
    \includegraphics[width=\linewidth,height=.13\textheight]{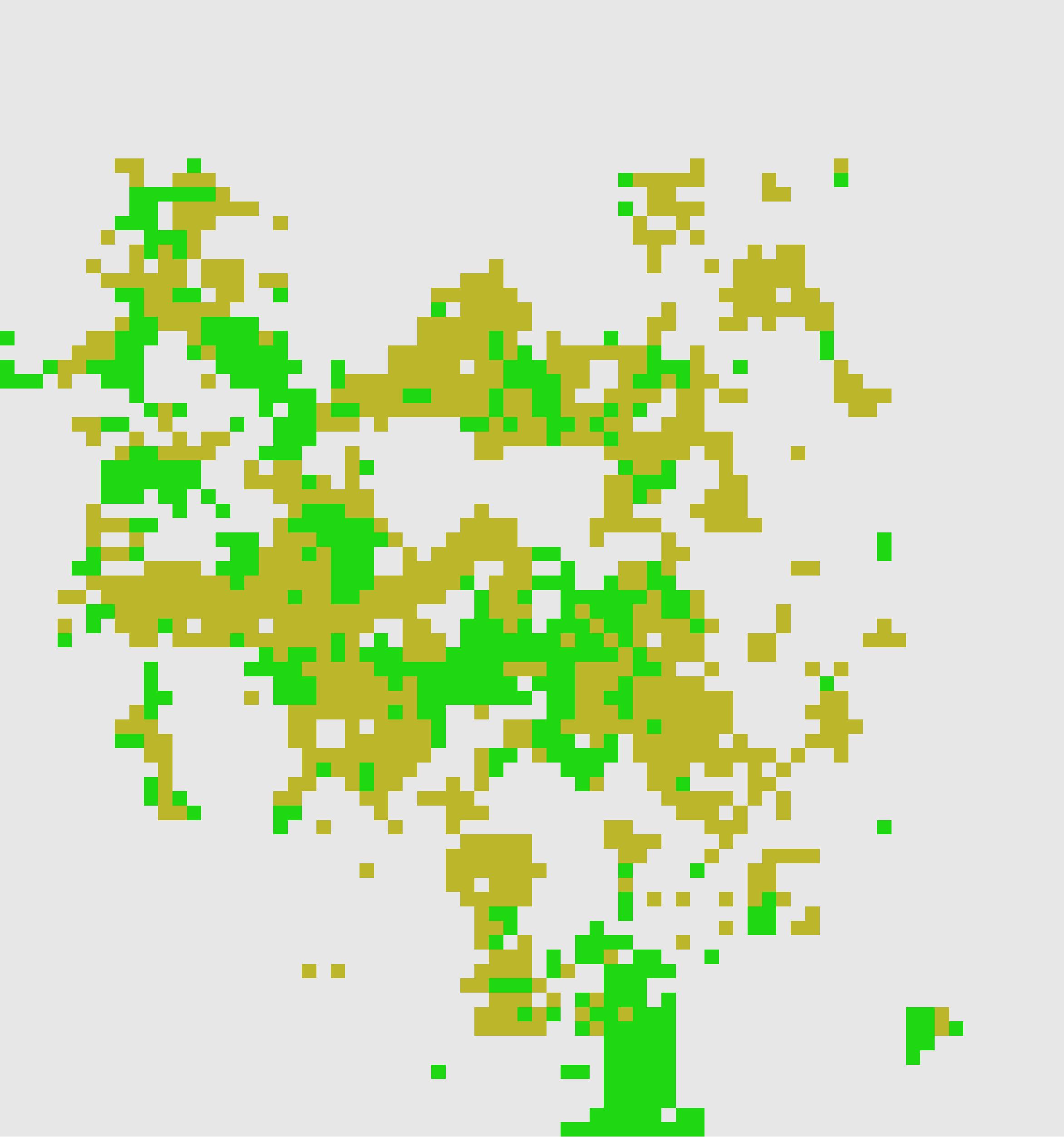}
    \caption{GV Occupancy.}
    \label{fig:gv}
\end{subfigure}

&
\begin{subfigure}{0.17\textwidth}
    \includegraphics[width=\linewidth,height=.13\textheight]{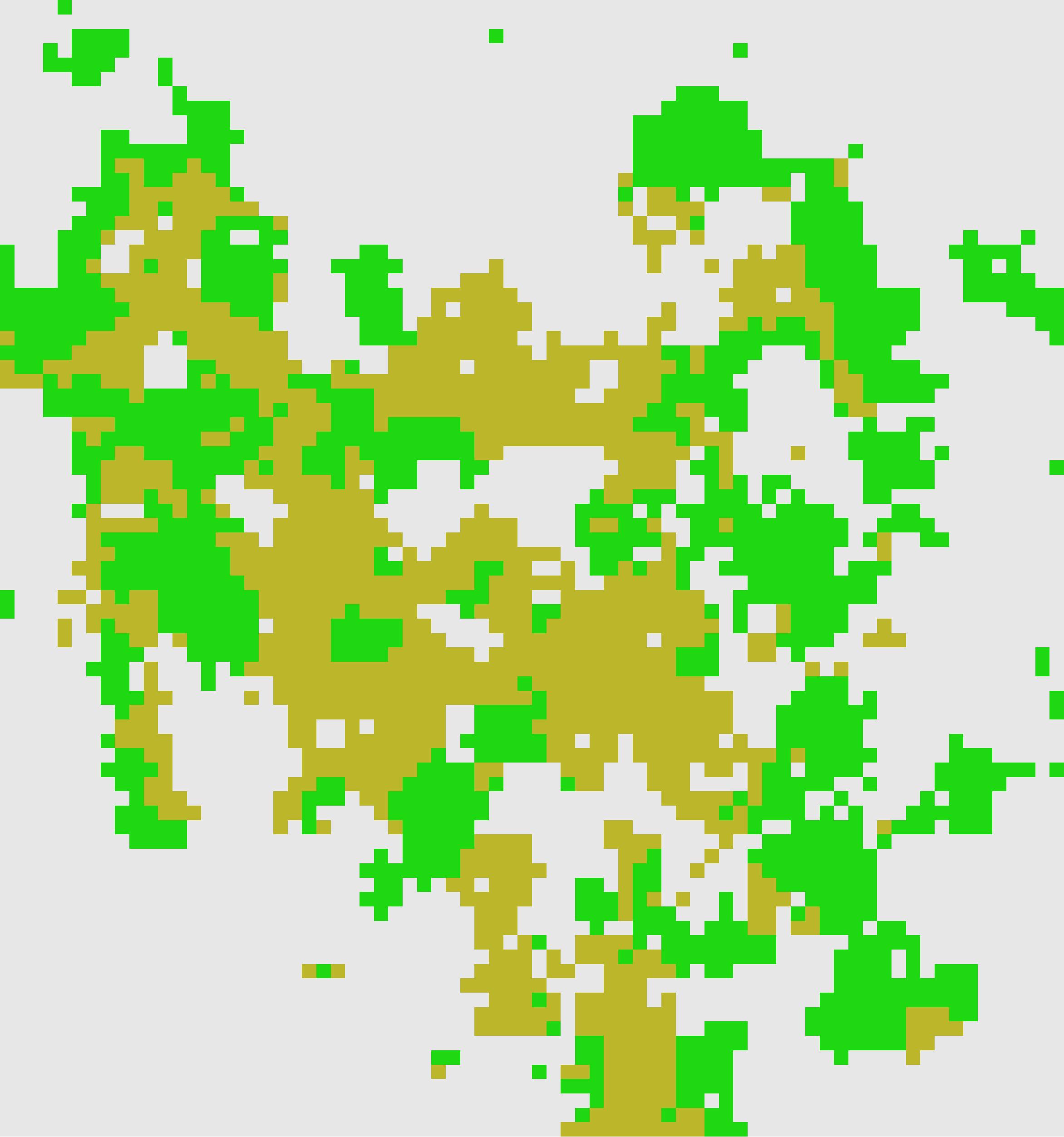}
    \caption{Understory Occupancy.}
    \label{fig:under}
\end{subfigure}

&
\begin{subfigure}{0.17\textwidth}
    %\scalebar{\includegraphics[width=\linewidth,height=.13\textheight]{images/raster_canopy_binary_gt16.jpeg}}{370}{10}{10} 
    \begin{tikzpicture}
    \node[anchor=south west,inner sep=0] (image) { 
    \includegraphics[width=\linewidth,height=.13\textheight]{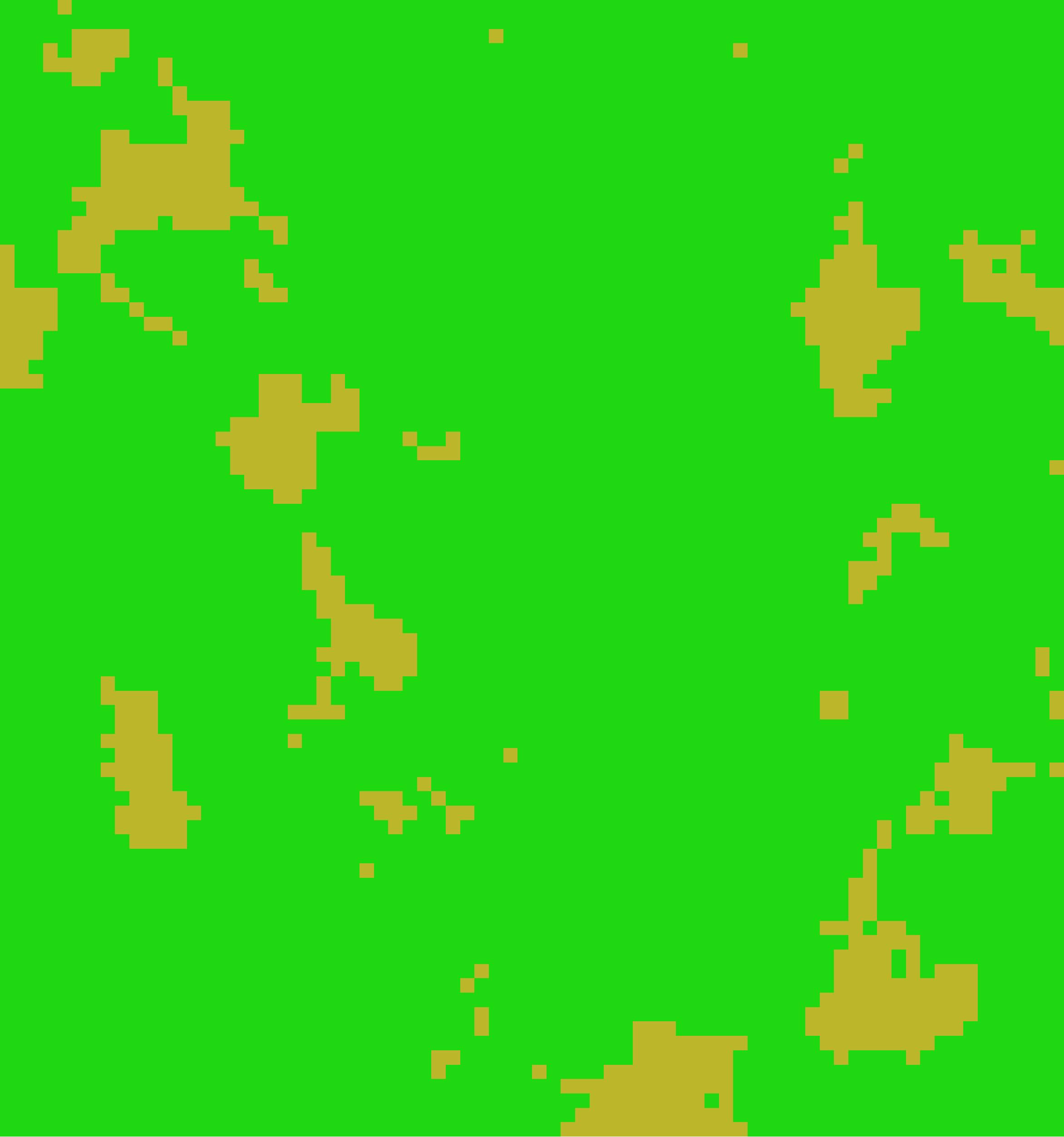}
    };
      \begin{scope}[x={(image.south east)},y={(image.north west)}]
    \draw [white, line width=0.2em] (0.04,1.2em) -- node[below,inner sep=0.1em, font=\small] {\SI{10}{\meter}} (.31,1.2em);
  \end{scope}
     \end{tikzpicture}
    \caption{Overstory Occupancy.}
    \label{fig:over}
\end{subfigure}
&
%\fbox{
\begin{minipage}[b][0.05\textheight][b]{0.1\textwidth}
\hspace{-4mm}
\begin{tabular}{@{}rl}
       \tikz \fill[fill=green, scale=0.3, draw=black] (0,0) rectangle (1,1); 
   & full\\
      \tikz \fill[fill=brown!50!yellow, scale=0.3, draw=black] (0,0) rectangle (1,1); 
   & empty\\
      \tikz \fill[fill=gray!20!white, scale=0.3, draw=black] (0,0) rectangle (1,1); 
   & no data\\
\end{tabular}
\end{minipage}
%}
\end{tabular}
\caption{\textbf{Annotations of WildForest3D.} 
Our dataset contains $29$ annotated dense forest plot, each with dense 3D annotations \subref{fig:3D_annot},
and ground truth occupancy maps at $50$cm resolution for the layers of ground vegetation \subref{fig:gv}, understory \subref{fig:under}, and overstory \subref{fig:over}. The colormap of \subref{fig:3D_annot} is the same as in Figure~\ref{fig:teaser}, with unannotated points in grey.}
\label{fig:our_dataset}
\end{figure*}

\paragraph{\bf Deep Learning-Based Vegetation Structure Analysis.}
Bolstered by their wide adoption by the computer vision and remote sensing communities, 3D deep learning methods \cite{guo2020deep} have started to be used for forestry analysis.
Lang \etal \cite{lang2022global} estimate the canopy height at global scale from satellite LiDAR data.
Ayrey \etal \cite{rs13245113} use a 3D CNN model to extract forest inventory indicators at $10$m resolution from voxelized ALS point clouds.
Liu \etal \cite{9300163} propose a deep-learning based regression model for estimating forest structural parameters from handcrafted 3D features.
Kalinicheva \etal \cite{kalinicheva2022predicting} use plot-level annotations to generate high resolution layer occupancy maps from ALS data. 
In this paper, we propose alternatively to explore the potential of using fine-grained 3D annotations to model the vegetation multi-layer structure using a deep network.

\section{WildForest3D}
%-------------------------------------------------------------------------
\label{sec:data}
In this section, we introduce WildForest3D, a novel dataset of aerial LiDAR scans of dense forests with dense 3D annotations and associated layer occupancy maps.

%.....................................
\subsection{Vegetation layers}
%.....................................

Relevant definitions of vegetation layers depend on the study area and are often subjective.
In this paper, we choose vegetation layers deemed fit by forestry experts after the analysis of the study area. However, the number and heights of the layers can easily be changed in our provided dataset generation scripts for specific biomes. In the rest of the paper, we operate with three layers: (i) Ground Vegetation (GV) comprises shrubs, small trees, ferns and other vegetation that are between $0.5$ and $1.5$m of height; (ii) understory refers to small trees or large shrubs vegetation with a height between $1.5$m and $5$m; (iii) overstory corresponds specifically to the crown of trees with a height above $5$m.
%.....................................
\subsection{Characteristics of WildForest3D}
%.....................................
\label{sec:data:3d}
The study area is located in the heavily forested French region of Aquitaine, and was scanned using a LiDAR installed on unmanned aerial vehicle with an average of $60$ pulses per square meter. Each point is associated with its coordinates in Lambert-93 projection, the intensity value of returned laser signal, and the echo return number. The elevation of the points is given using a digital elevation model, such that the ground points are always at $z=0m$.

Forestry experts selected $29$ representative study plots of size around $40\times40 m^2$, for a total of $47\,000 m^2$. Airborne LiDAR data of dense forest cannot be annotated without \textit{in situ} data collection, as the branches of trees and shrubs intersect and are too difficult to parse from just 3D point clouds. This makes the annotation extremely costly and error-prone, all the more that the plots are typically not easily accessible.
In each plot, an operator takes detailed measurements of the trees in its immediate proximity: position, height, inclination, species, stem circumference, and height of crown base. This data is reported in polar coordinates relative to the plot center, which is not directly conducive to supervising machine learning algorithms operating on 3D points.

\subsection{3D Point Annotation.} We designed a semi-automatic approach to convert these annotations into point labels across a set $\mathcal{C}$ of $6$ classes: 
\emph{ground}, \emph{ground vegetation} (GV), \emph{understory}, tree \emph{stem}, and
\emph{coniferous} and \emph{deciduous} tree crowns. 
As seen in Figure~\ref{fig:stratum_example}, each point is classified according to the vegetation instance it belongs to, and not its elevation. A point at a height of $3$m but belonging to a coniferous with a tree top height of $8$m will be classified as coniferous tree crown, and not as understory. Due to its nature, ground vegetation is particularly hard to annotate in 3D and is usually only reported by experts through local coverage ratio \cite{WING2012730}. In our dataset, we chose to remove the few ground vegetation label altogether for 3D points, as they proved inconsistent. However, as we show in the next section, we can still recover this important layer through a geometric analysis and using other annotations.

\begin{figure*}[t]
  \centering
%   \fbox{\rule{0pt}{2in} \rule{0.9\linewidth}{0pt}}
  \includegraphics[width=1\linewidth, trim=0cm 0cm 1.15cm 0cm,clip]{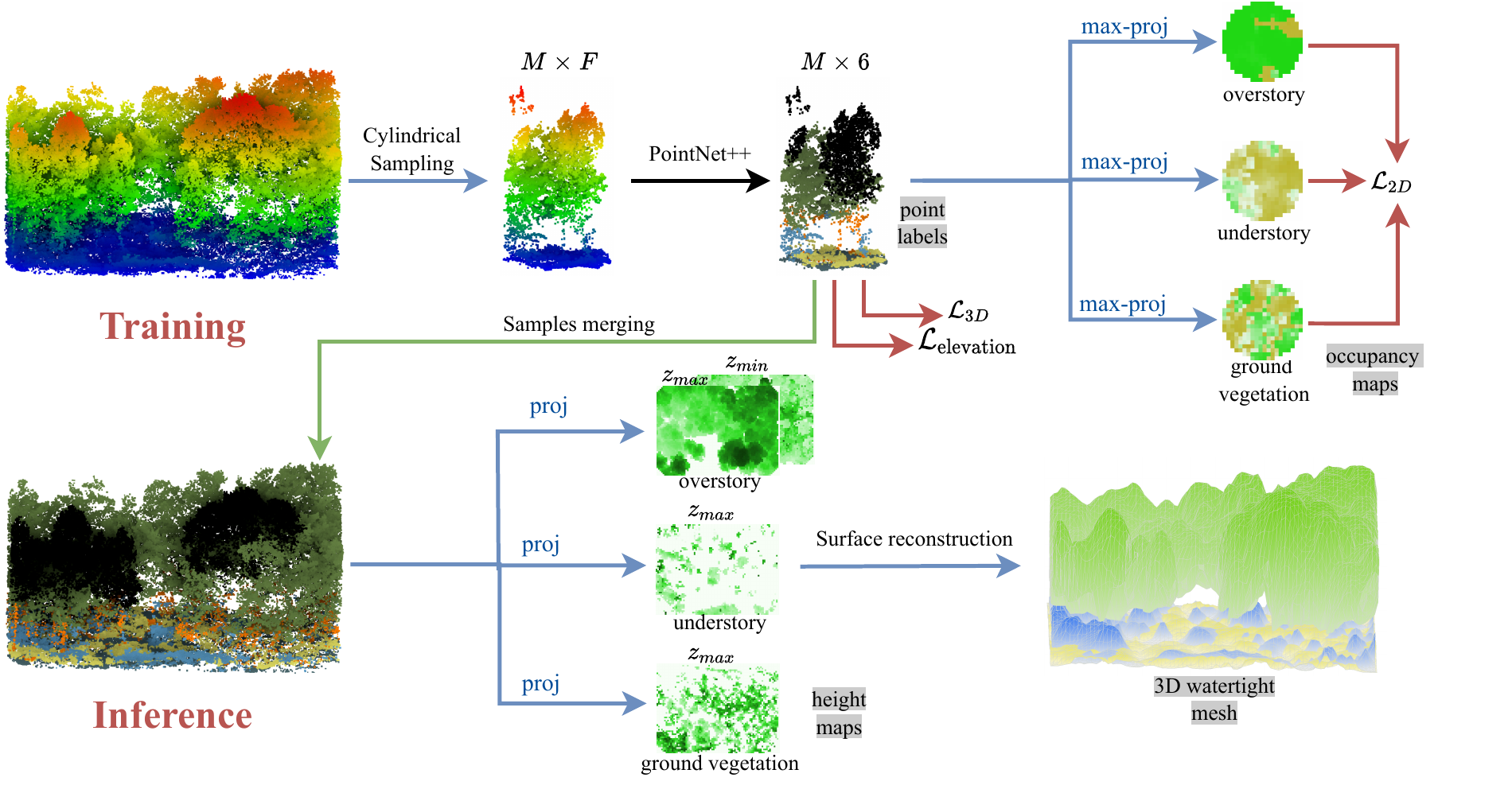}
  \caption{\textbf{Pipeline.} During  training, our network performs the semantic segmentation of a 3D point cloud sample within $6$ different classes. The point probabilities are projected onto rasters to obtain soft occupancy maps for $3$ different vegetation layers. The network is supervised using both 2D and 3D predictions. During inference, the predictions are computed for an entire plot and the prediction are used to derive the minimum and maximum elevation of each layer. Finally, we can produce a watertight 3D mesh representing each vegetation layer.}
  \label{fig:model}
\end{figure*}

%........................................
\subsection{Layer Occupancy Maps}
\label{sec:2d_dataset}
%.........................................
Since our goal is to model the multi-layer nature of dense forests, we generate binary occupancy maps for the ground vegetation ${O}^\text{GV}$, understory ${O}^\text{under}$, and overstory ${O}^\text{over}$, see Figure~\ref{fig:our_dataset}. 
Our maps are of size $H\times W$, corresponding to an adjustable resolution of $r=0.5\:$m per pixel. Each pixel of each map is associated with either the value \emph{full}, \emph{empty}, or \emph{no data}. 
This last value represents the ambiguity induced by partial annotations: not all trees and bushes are reported, and no ground vegetation label is annotated.
%This last value represents the ambiguity induced by partial annotations and the absence of ground vegetation point labels.

The occupancy maps are obtained by vertically projecting the 3D points to their corresponding stratum. First, we project all points with labels \textit{deciduous} and \textit{coniferous} to the overstory layer to determine the full pixels of this map. We repeat this process for the understory by projecting points with the understory label. 
We do not have explicit labels for the ground vegetation and the annotations are partial. However, we can use the elevation to determine the ground  vegetation layer map and complete the understory and overstory layers. However, using unannotated points yields more ambiguity, which we represent with the no data class.

First, we remove all annotated points. 
For each pixel, we compute the maximal height of all remaining points in its vertical extent. We then annotate the ground vegetation map and the pixels of the understory and overstory maps \emph{that are not yet marked as full} according to this height:
\begin{itemize}
    \item under $0.5$m: the pixels of all maps are marked as empty;\vspace{-0.15cm}
    \item  between $0.5$ and $1.5$m: the ground vegetation pixel is annotated as full, and the pixels of both overstory and understory are marked as empty;\vspace{-0.15cm}
    \item between $1.5$ and $5$m: the pixels of the overstory are marked as empty, the understory as full, and the ground vegetation as no data.\vspace{-0.15cm}
    \item over $5$m: the pixels of the overstory are full, and the other maps as no data.
\end{itemize}
This geometric criterion does not overwrite full pixels of the overstory and understory derived from explicit point labels. The canopy layer do not have pixels without label, except if no points fall into the pixel. 
\section{Methodology}
 
Our objective is to automatically derive from an aerial 3D LiDAR scans the occupancy and depth of different vegetation layers: ground vegetation, understory, and overstory. As illustrated in Figure~\ref{fig:model}, we compute 3D point classification and 2D occupancy maps, which are both used for supervision. Once trained, our approach can be used to derive the layers' thickness, and even to produce watertight meshes for each layer of such large scans.
%....................................
\subsection{3D Point Classification}
\label{sec:point_cl}
%....................................
%Each plot is presented as a point cloud of $N$ points 
During training, we split the data into batches comprising $N$ cylindrical samples of radius $R$ and infinite vertical extent, randomly sampled on a grid with step size $R/5$. Each point of the cylinders is associated with a vector of $F$ features containing their position, radiometric information, and the number of echoes. The point's horizontal coordinates are given relative to the center of the cylinder. For the rest of the section, we consider a cylinder of $M$ points and denote its point feature as $x \in \mathbf{R}^{M \times F}$.

This cylinder is randomly subsampled to a fixed number of points $S$, which are then classified using the PointNet++ (PTN++) network \cite{pointnet2} among the class set $\mathcal{C}$ defined in Section~\ref{sec:data:3d}. The prediction is then upsampled to the full resolution point cloud with nearest neighbor interpolation to obtain a predicted probability vector $p \in [0,1]^{M \times 6}$:
\begin{align}
    p = \upsample_{S \mapsto M} \circ\, \text{PTN++} \circ \dosample_{M \mapsto S }(x)~.
\end{align}
The point-wise predictions are supervised with the cross-entropy loss $\mathcal{L}_\text{3D}$, using the point annotations.
%.................................
\subsection{Layer Occupancy Prediction}
\label{sec:point_proj}
%.................................
We now predict occupancy maps of size $[H,W]$ and pixel size $s$ for each layer $l$ in $\{\text{GV}, \text{understory}, \text{overstory} \}$. For $i,j \in [1\cdots H,1 \cdots W]$, we denote by $\proj(i,j)$ the set of 3D points of the considered cylinder whose vertical projection falls into the pixel $(i,j)$. We define the occupancy $o^l_{i,j}$ as the maximum probability associated with  the points of the relevant class:
\begin{align}
    o^\text{GV}_{i,j} &= \max_{k \in \proj(i,j)} p^\text{GV}_k~, \\
    o^\text{understory}_{i,j} &= \max_{k \in \proj(i,j)} p^\text{understory}_k~, \\
    o^\text{overstory}_{i,j} &= \max_{k \in \proj(i,j)} \left(p^\text{deciduous}_k + p^\text{coniferous}_k\right)~.
\end{align}
Note that neither the stem class nor the ground class are projected onto rasters, as they do not participate in the vegetation coverage. The predicted occupancy maps are supervised with the ground truth occupancy maps described in Section~\ref{sec:2d_dataset}, and using the binary cross-entropy loss $\mathcal{L}_\text{2D}$ applied pixel-wise.
%..........................................
\subsection{Unsupervised Elevation Modeling}
\label{sec:elevation_modeling}
%..........................................
% \begin{figure}
%     \centering
%     \begin{tikzpicture}[scale=1]
%     \begin{axis}
%     [xmin=0,xmax=30,height=6cm,width=1\linewidth,
% 	 ymin=0,ymax=0.2, ytick={0.1,0.2},
% 	 xlabel=height (m), ylabel=probability density, ybar,legend cell align={left},axis lines=left]
%     \addplot [ybar, bar width=2pt, color=blue,fill=blue] table[col sep=comma,header=false, x index = 0,y index=1] {./images/ECM.csv};
%     \addlegendentry{\small{Empirical distribution}}
%     \addplot [sharp plot,color=green!80!black, ultra thick,line legend] table[col sep=comma,header=false, x index = 0,y index=2] {./images/ECM.csv};
%     \addlegendentry{\small{Lower stratum $\Gamma^\text{lower}$}}
%     \addplot [sharp plot,color=red, ultra thick,line legend] table[col sep=comma,header=false, x index = 0,y index=3] {./images/ECM.csv};
%     \addlegendentry{\small{Higher stratum: $\Gamma^\text{higher}$}}
%      \end{axis}
%   \end{tikzpicture}
%     \caption{{\bf Elevation Modeling.} The elevation across the entire dataset can be modeled by a mixture of Gamma distributions. }
%     \label{fig:emc}
% \end{figure}
To overcome the problem of partial annotations, we propose to explicitly model the elevation of points within two aggregated strata as suggested by Kalinicheva \etal ~\cite{kalinicheva2022predicting}.
We model the distribution of the point elevations as a mixture of two Gamma distributions: the lower (ground and ground vegetation) and higher strata. This mixture can be approximated using the Expectation–Conditional–Maximization algorithm \cite{young2019finite} with manual initialization but without explicit stratum supervision.% (Figure~\ref{fig:emc}).

The learned distributions $\Gamma^\text{lower}$ and  $\Gamma^\text{higher}$ of the lower and higher strata allow us to define an additional loss as the average negative log-likelihood of the elevation of all points:
\begin{align}\nonumber
    \mathcal{L}_\text{elevation}
    &=-\frac1M \sum_{m} \log\left((p^\text{soil}_m + p^\text{GV}_m) \Gamma^\text{lower}(z_m)\right.\\
    & \!\!\!\!\!\!\!\!\!\!\!\!\!\!\!\!\!\!\!\!\!\!\!\!\left.  \qquad    + (p^\text{under}_m + p^\text{decid}_m + p^\text{conif}_m + p^\text{stem}_m) \Gamma^\text{higher}(z_m) \right)~.
\end{align}
This loss encourages points with low elevation (high $\Gamma^\text{lower}$) to be predicted as ground or GV, and conversely for $\Gamma^\text{higher}$.
%...............................
\subsection{Model supervision}
\label{sec:loss_functions}
%...............................
We supervise our model by combining three different losses: point-wise, pixel-wise, and elevation:
\begin{align}\label{eq:loss}
    \mathcal{L} = \mathcal{L}_\text{3D} +  \lambda\mathcal{L}_\text{2D} + \mu \mathcal{L}_\text{elevation}~,
\end{align}
with $\lambda,\mu$ hyperparameters, set to $1$ and $0.1$. We train all models for $100$ epochs of $N=1000$ cylinder with a batch size of $5$, a weight decay of $\footnotesize{10}^{-3}$, and a learning rate of $5\cdot\scriptsize{10^{-4}}$, divided by $2$ every $20$ epochs. We use data augmentation by randomly rotating cylinder point clouds around the $z$-axis and adding a random Gaussian noise $\mathcal{N}(0,0.01)$ clipped to $[-0.03, 0.03]$ to the normalized intensity values.

%..................................
\subsection{Inference}
\label{sec:reconstruction}
%..................................
Once fully trained, our model can be used to model the vegetation layers on entire scanned areas. We sample cylinders on a grid of size $R$ and compute the prediction for each point. When cylinders overlap, we average the predicted probabilities of points appearing in several cylinders.
We associate to each point a hard assignment $l \in \mathcal{C}^M$ as the class with the highest predicted probability. This allows to define for each layer a binary occupancy map $\hat{o}^l$: a pixel is full if at least one point of the corresponding classes falls in its vertical extent. We can also define the minimal and maximal heights of each layer by considering the points of $\proj(i,j)$ with corresponding labels. We set the lower boundary of the GV and understory layers as $0$.

Using the occupancy, the minimal and maximal height maps, we can reconstruct a watertight triangular mesh for each layer by connecting adjacent non-empty pixels in a $4$-neighborhood graph and using their corresponding height. Such side-product mesh can be used for downstream visualisation, computational or simulation tasks such as biomass estimation, forest fire prevention, ecological studies or ground illumination modeling \cite{Fire_simu,Amazonian_strata}.

%-------------------------------------------------------------------------

%As in training step, we start by computing point-wise predictions on the grid-subsampled cylinders. Contrary to the training, we use all the cylinders sampled on the regular grid with step $G_i = R$ to assure the whole plot coverage. The cylinders are merged afterwards to reconstruct the whole point cloud. The subsampled clouds belonging to one plot overlap with each other, so one point may be associated to several prediction probability values. Thus,we compute the overlapping point final class predictions by applying the $argmax$ function to the average of the class probabilities.
%Finally, each class is projected to the corresponding vegetation raster layer.

%To compute the thickness of each layer, we produce 2.5D occupancy maps with upper and lower vegetation boundaries. To produce the upper boundary elevation map, we associate to each vegetation pixel the maximum $z$ value of all the points that fall into this pixel. By default, the lower boundaries of shrub and understory layers are $0m$. We obtain the lower boundary of the canopy layer in the similar way by projecting the minimum $z$ values to the vegetation-filled pixels. Finally, we construct a mesh representing each vegetation layer by connecting the neighbouring pixels to each other.

%--------------------------------
\section{Numerical Experiments}
%----------------------------------------

\begin{figure*}
\centering
\begin{tabular}{@{}cccc}

\begin{subfigure}{0.24\textwidth}
    \includegraphics[width=\linewidth]{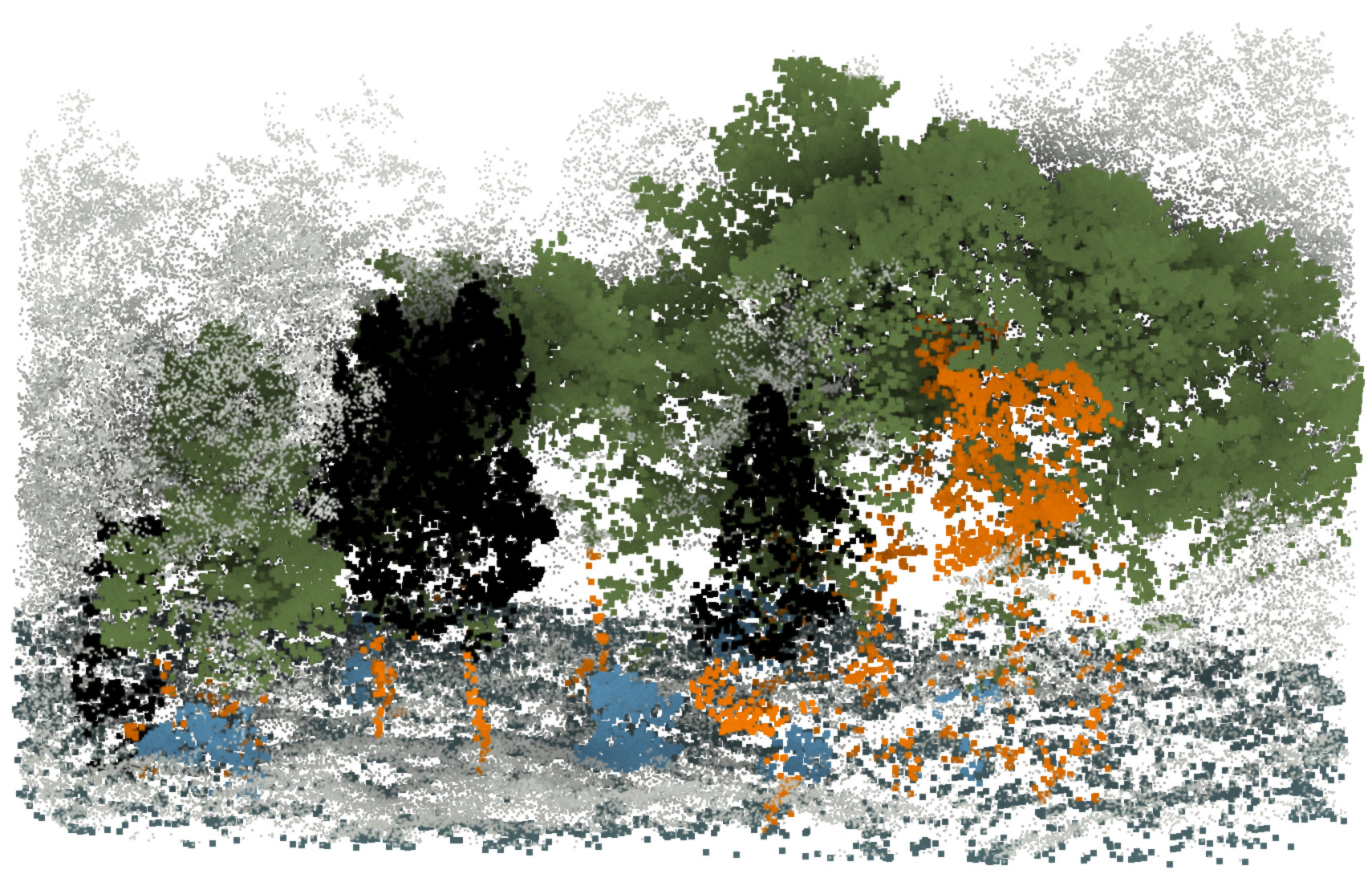}
    \caption{True Point Labels.}
    \label{fig:gt_points15}
\end{subfigure}
&
\begin{subfigure}{0.22\textwidth}
    \includegraphics[width=\linewidth]{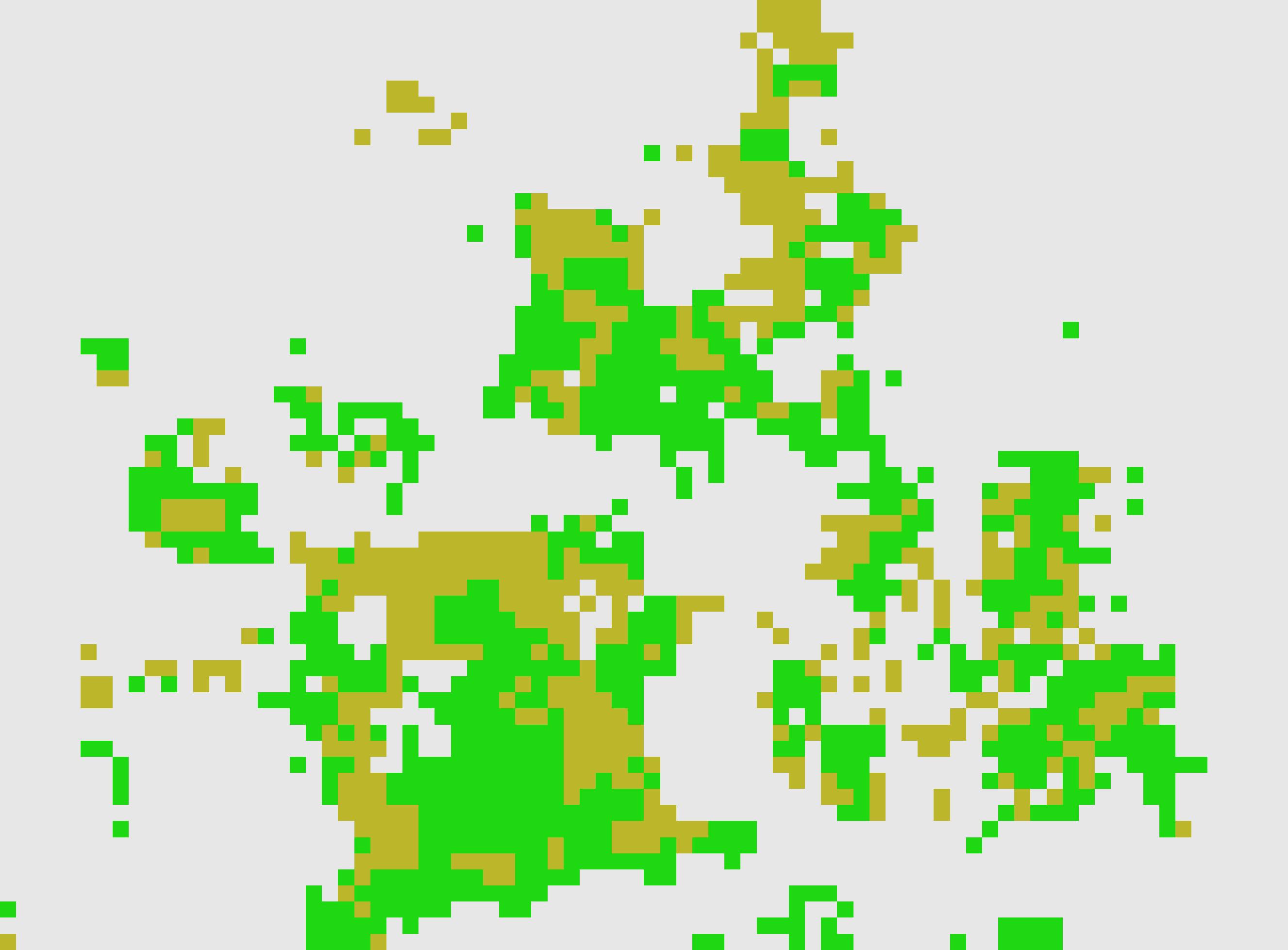}
    \caption{True GV Occupancy.}
    \label{fig:gv15}
\end{subfigure}

&
\begin{subfigure}{0.22\textwidth}
    \includegraphics[width=\linewidth]{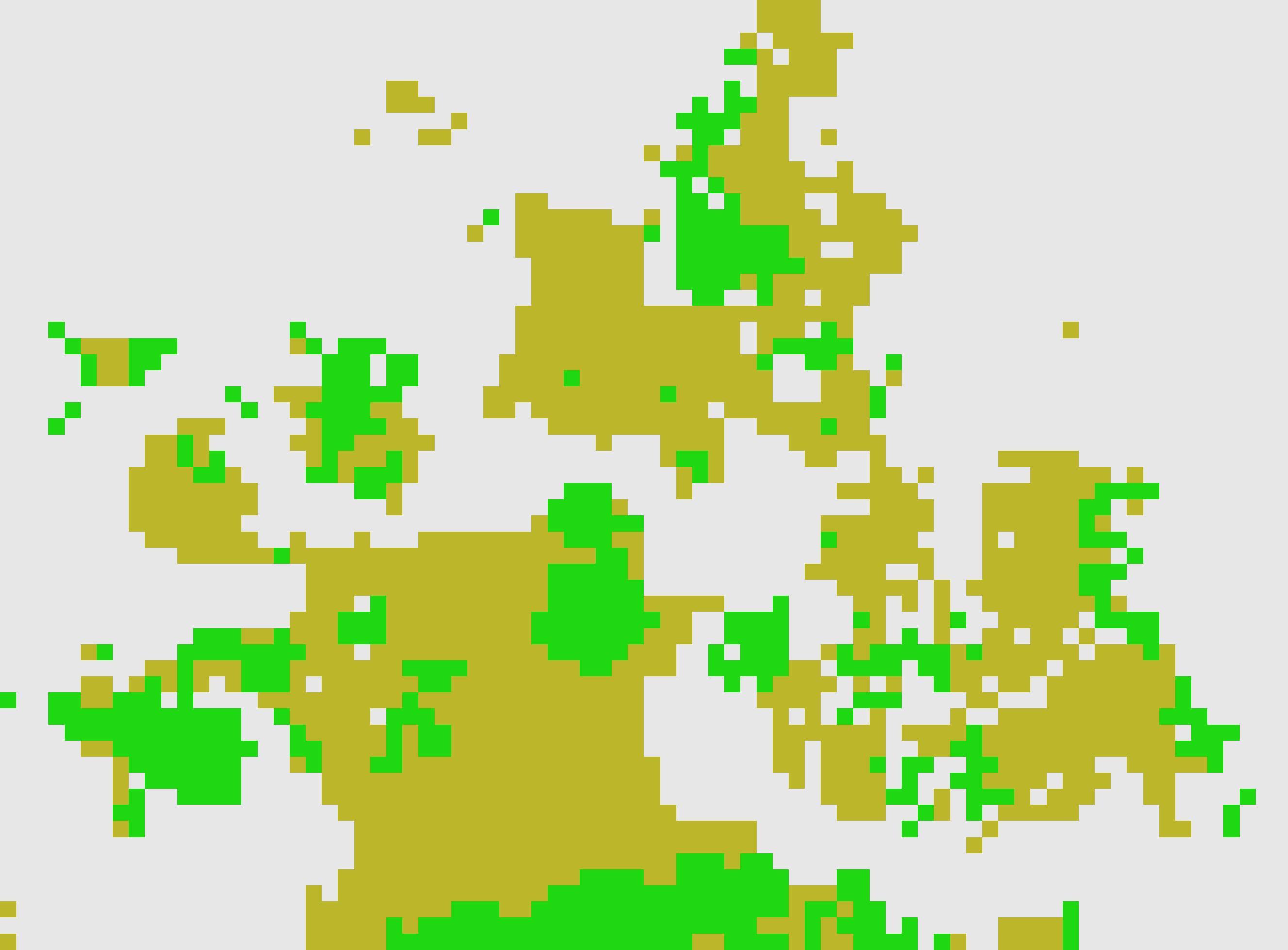}
    \caption{True Understory Occupancy.}
    \label{fig:under15}
\end{subfigure}

&
\begin{subfigure}{0.22\textwidth}
    \begin{tikzpicture}
    \node[anchor=south west,inner sep=0] (image) { 
    \includegraphics[width=\linewidth]{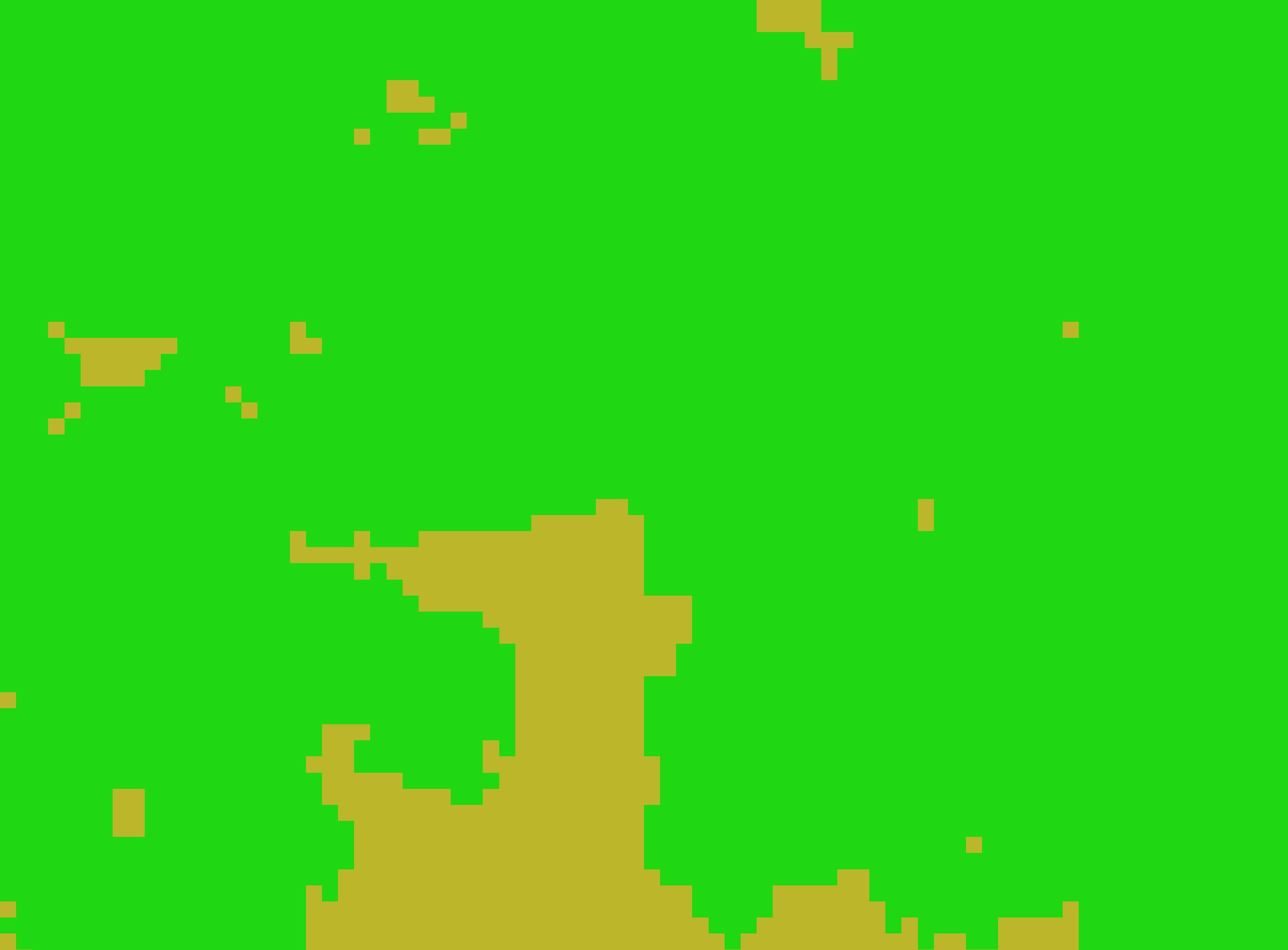}
    };
      \begin{scope}[x={(image.south east)},y={(image.north west)}]
    \draw [white, line width=0.2em] (0.04,1.2em) -- node[below,inner sep=0.1em, font=\small] {\SI{10}{\meter}} (.29,1.2em);
  \end{scope}
     \end{tikzpicture}

    %\scalebar{\includegraphics[width=\linewidth]{images/raster_canopy_binary_gt15_clip.jpeg}}{400}{10}{10} 
    % \includegraphics[width=\linewidth]{images/raster_canopy_binary_gt15_clip.jpeg}
    \caption{True Overstory Occupancy.}
    \label{fig:over15}
\end{subfigure}
\\

\begin{subfigure}{0.24\textwidth}
    \includegraphics[width=\linewidth]{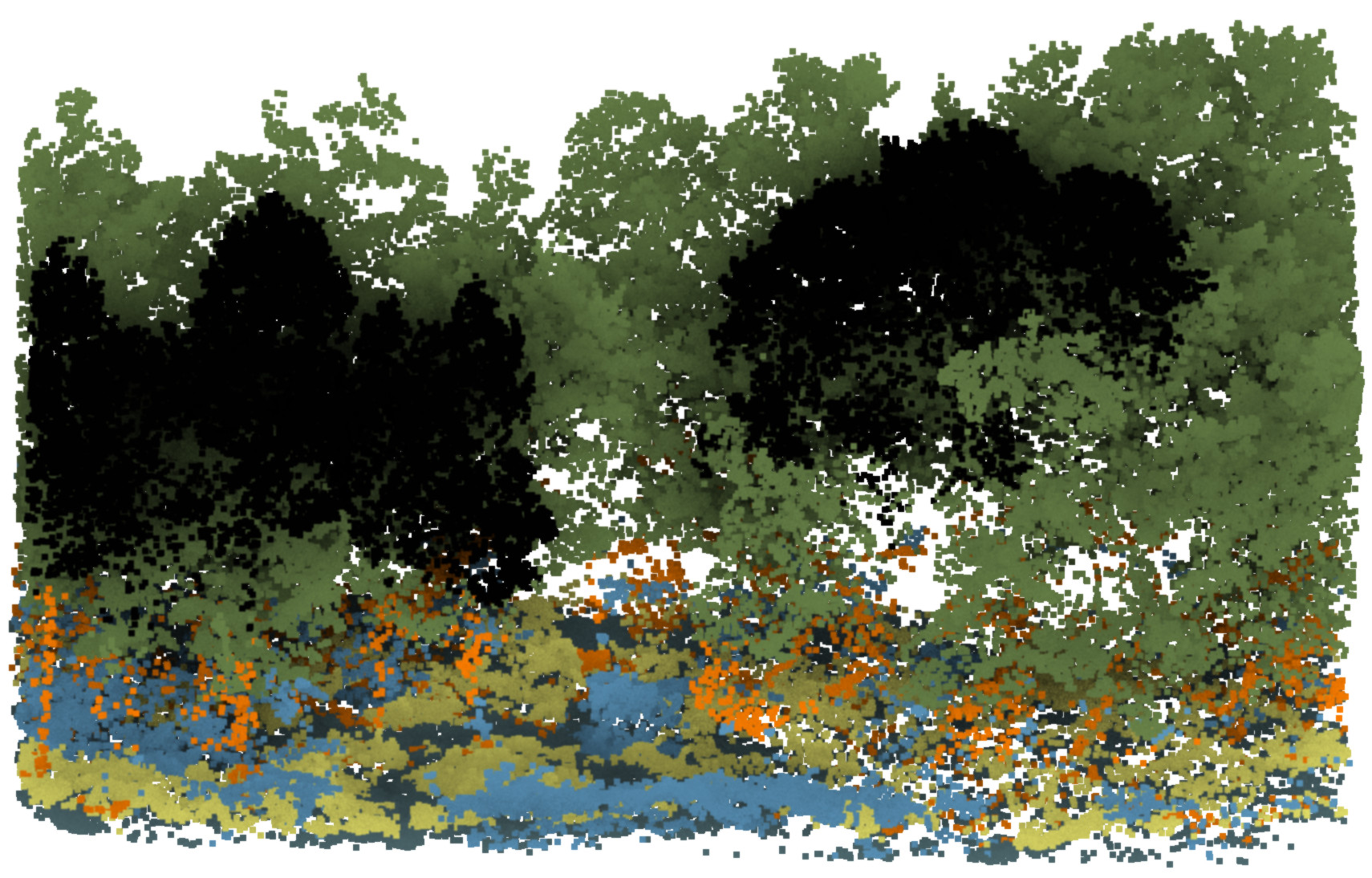}
    \caption{Point-wise Prediction.}
    \label{fig:pred_gt_points15}
\end{subfigure}
&
\begin{subfigure}{0.22\textwidth}
    \includegraphics[width=\linewidth]{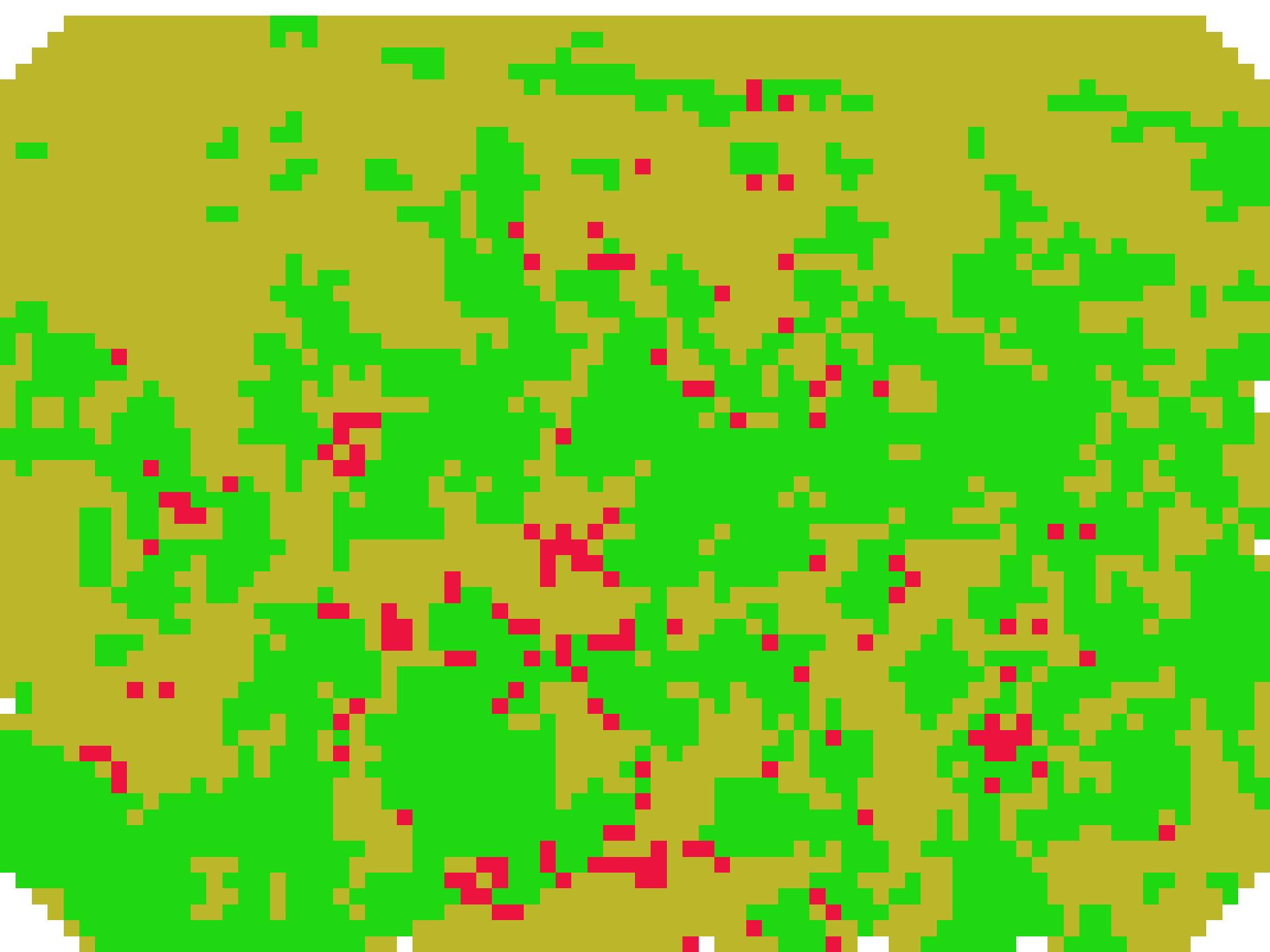}
    \caption{Pred. GV Occupancy.}
    \label{fig:pred_gv15}
\end{subfigure}

&
\begin{subfigure}{0.22\textwidth}
    \includegraphics[width=\linewidth]{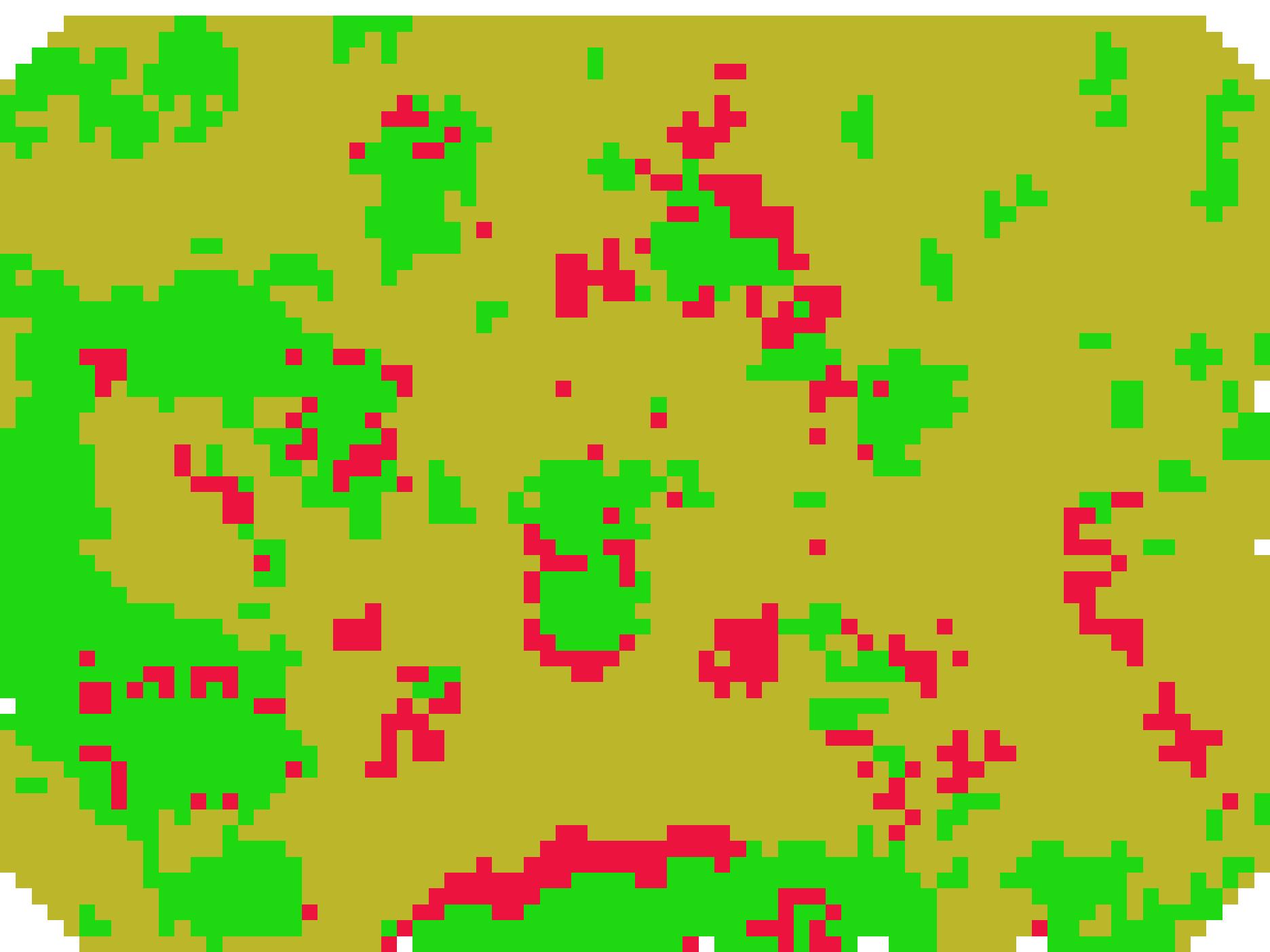}
    \caption{Pred. Understory Occupancy.}
    \label{fig:pred_under15}
\end{subfigure}

&
\begin{subfigure}{0.22\textwidth}
    \includegraphics[width=\linewidth]{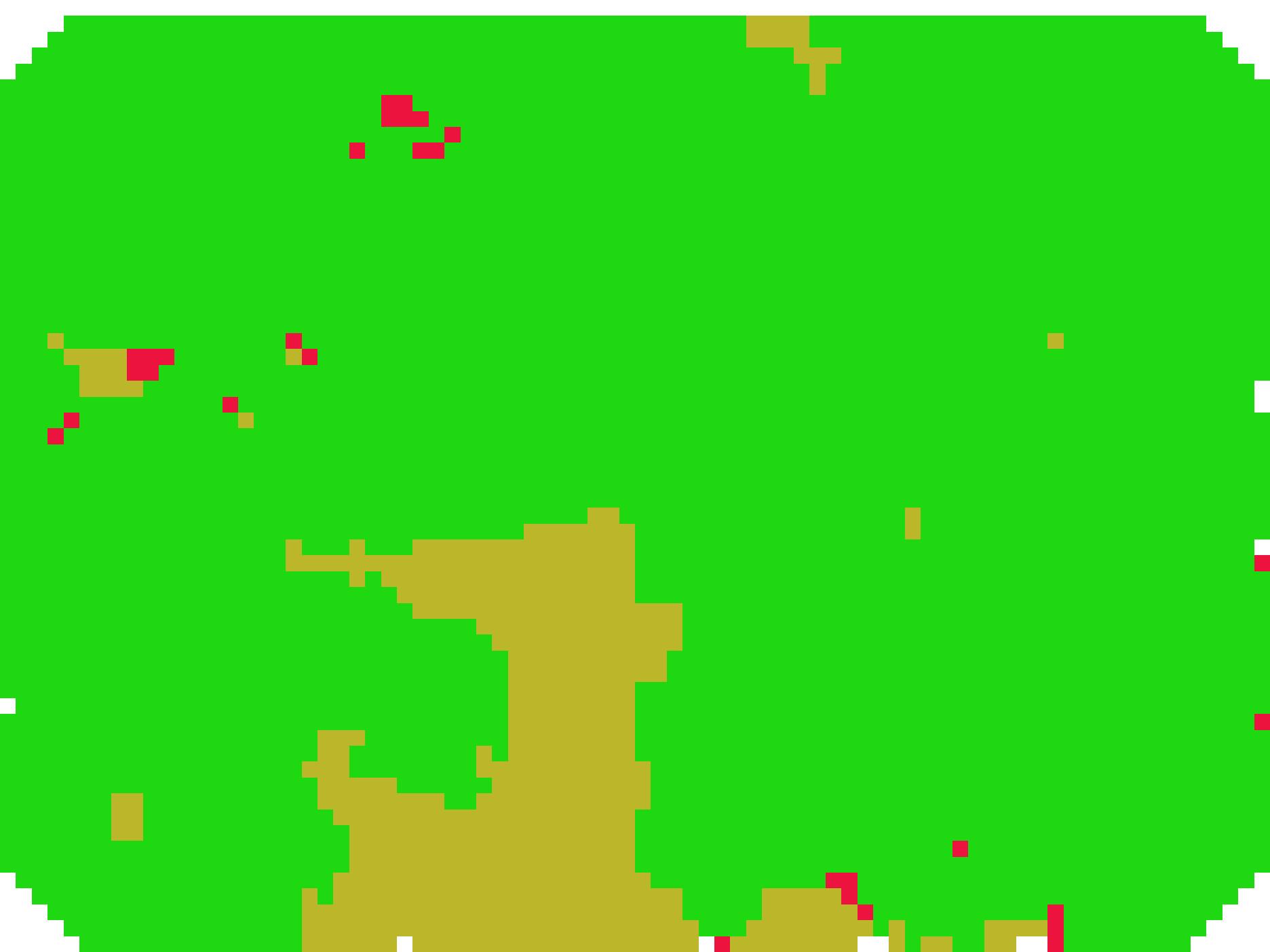}
    \caption{Pred. Overstory Occupancy.}
    \label{fig:pred_over15}
\end{subfigure}
\end{tabular}
\caption{\textbf{Qualitative Results.} We report the ground truth (top row) and prediction (bottom row) for the point labels and layer occupancy maps. {The prediction errors are in \textcolor{red}{red}}.}
\label{fig:results_plot15}
\end{figure*}

In this section, we present quantitative and qualitative evaluations of our approach. To the best of our knowledge, no method exists with available implementation to derive fine-grained height maps across different vegetation layers from airborne LiDAR point clouds. To encourage reproducibility, both our dataset and method are released in open-source.
%..............................
\subsection{Experimental settings}
%..............................
We select the hyperparameters of our model to retain a fast inference and high precision on a NVIDIA RTX 3060 GPU.
We use a pixel size of $0.5\:$m for the occupancy and height raster, and a sample cylinder radius of $R=5\:$m.
To handle the varying density of point clouds and to facilitate batch processing, we randomly sample  $S=2^{14}=16\,384$ points in each cylinder. %As the density of the lower layers is typically lower than the canopy, lower density leads to severe under-sampling. 

We base our implementation on PyTorch Geometric's  \cite{Fey/Lenssen/2019} version of PointNet++ \cite{pointnet2}. Our model configuration has a total of $147\,638$ parameters, see our repository for details. Our model can be trained from scratch in $7\:$hours, and can process $75\:$m$^2$/second in inference on a standard workstation.

%The following architecture is deployed for our PointNet++ model for semantic segmentation (see \cite{pointnet2} for more details): $SA(12288, 0.1, 0.33, [16, 32]) \rightarrow SA(4096, 0.25, 0.5[32,64]) \rightarrow SA(1024, 0.5, 0.5[64,128]) \rightarrow SA([128,128]) \rightarrow FP(128, 128) \rightarrow FP(128, 64) \rightarrow FP(64, 32) \rightarrow FP(32, 32, 32, 32, C)$, where $SA(K,r,a[l_1, ..., l_d])$ is a set abstraction (SA) level with $K$ local regions of ball radius $r$ and $a$ ratio of selected points using PointNet of $d$ fully connected layers with width $l_i (i = 1, ..., d)$. $SA([l_1, ...l_d])$ is a global set abstraction level that converts set to a single vector.

%We set the learning rate to $lr=0.0005$ with weight decay $wd=0.5$ each 20 steps. The Adam algorithm is used for model optimization.

%All the codes are developed in $Python$ using $PyTorch$ library for deep learning. Our network can be trained in 7 hours on an NVIDIA GeForce RTX 3060 GPU and a Xeon W-2123 CPU with $64$GB of RAM. The inference step takes BLABLA SECONDS per plot (BLABLA per m²).
%.............................
\subsection{Evaluation Metrics and Baselines}
%.............................

\begin{table*}
    \caption{{\bf 2D and 3D Prediction Performances:}  Intersection-over-Union and Overall Accuracy of the prediction of 3D point (3D) and 2D occupancy maps regression (2D) on the evaluation set. We evaluate our approach for the 3D task, as well as a regression and random forest-based baselines for the 2D tasks. Our method performs both tasks simultaneously.}
        \label{tab:3D_2D_results}
        \centering
        \begin{tabular}{llcccccccc}
            \hline
            &&\multicolumn{7}{c}{IoU, \%}&\multirow{2}{*}{OA, \%}\\
            \cline{3-9}
            Task & Method
                & Ground& Ground veg.   & Understory & Decid. & Conif. & Stem & Mean \\
            \hline
            3D & Ours &95.1   &-   & 43.3    & 90.0 & 23.5 & 15.5 & 53.5 & 90.5 \\\cline{6-7}
            2D & Ours & - & 81.5 &61.0  & \multicolumn{2}{c}{99.3} & - &80.6&92.3\\
            2D & Logistic Regression & - & 62.2 &40.5  & \multicolumn{2}{c}{99.5} & - &67.4 & 87.3  \\
            2D &  RF Classifier & - & 89.3 &62.5  & \multicolumn{2}{c}{99.6} & - &83.8 & 94.2  \\\bottomrule

    \end{tabular}
\end{table*}

We evaluate the quality of the point prediction, occupancy maps, and height prediction with the following metrics:
\begin{itemize}
    \item {\bf 3D Prediction:} We use the Overall Accuracy (OA) and the Intersection-over-Union (IoU) of the point-wise predictions for all classes, except the ground vegetation class, which is not annotated in the 3D supervision. mIoU3D refers to the class-wise average over the classes without weight.\vspace{-0.15cm}
    \item {\bf 2D Raster Prediction:} We measure the performance with the OA and IoU of the vegetation binary occupancy prediction for all vegetation layers. mIoU2D is averaged across all three layers.\vspace{-0.15cm}
    \item {\bf Height Prediction:} We use the Mean Absolute Error (MAE) and Mean Relative Error (MRE) of the layer height predictions, independently computed for each layer and only for pixels with full occupancy.
   The MRE is defined as the ratio between the absolute value of the error and the true height of the considered pixel.
\end{itemize}

In order to put our results into perspective, we implement four simple baselines in the spirit of regression-based methods \cite{jarron2020detection, 10.1093/forestry/cpv032, 10.1371/journal.pone.0220096}. For each pixel, we compute handcrafted descriptors of the points above (maximum, minimum, mean, and standard deviation of $z$ values, mean intensity value, mean return number, and number of points in 10 irregular bins of elevation between $0$ and $30\:$m). We then train a logistic regression and a Random Forest (RF) classifier to predict 2D occupancy maps, and a linear regression and RF regression to predict height maps. Both RF models were set with maximum tree depth of $4$ and $100$ estimators.
%-------------------------------------------------------------------------
\subsection{Analysis}
%-------------------------------------------------------------------------
We select three plots for evaluation for their overall representativeness and the quality of their annotation. We remove them from the training set.
Please refer to Figure~\ref{fig:results_plot15} for visualisation of the results for one of those plots.

\begin{figure}
  \centering
%   \fbox{\rule{0pt}{2in} \rule{0.9\linewidth}{0pt}}
  \includegraphics[width=0.95\linewidth]{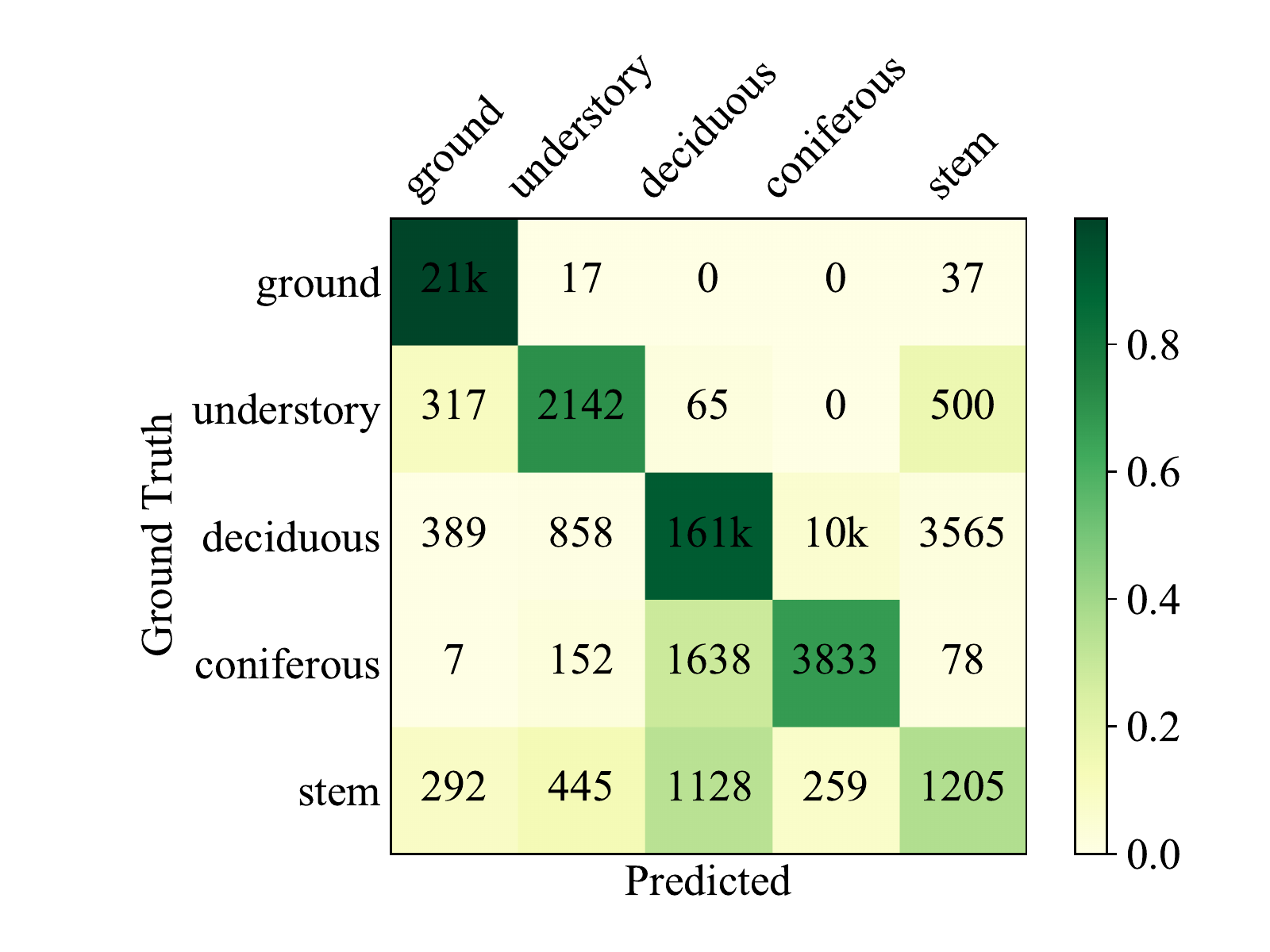}
  \caption{\textbf{3D Prediction Confusion Matrix.} Cells are colored according to the proportion of points with the corresponding class. We observe a strong class imbalance, and a concentration of errors between the deciduous, coniferous, and stem classes. }
   \label{fig:cm}
   \vspace{-3mm}
\end{figure}

In Table~\ref{tab:3D_2D_results}, we report the quantitative performance for 3D point classification and 2D binary vegetation occupancy maps. 
While we observe lower score for stems, this class is particularly hard to annotate and may not be as accurately labeled as the other classes.
We also notice the poor IoU score for the coniferous trees. This can be explained by the fact that the coniferous trees are more rare in our dataset that the deciduous ones and are taller in general. Hence, some tall deciduous trees are mistakenly classified as coniferous and vice versa, see Figure~\ref{fig:cm}.

The binary occupancy rasters are overall well predicted, with a lower score for the challenging understory layer.
The logistic regression and Random Forest baselines show good results for the prediction of the occupancy maps. However, note that our approach performs both 3D prediction and 2D regression simultaneously, which is necessary to recover the ground vegetation. Furthermore, the 2D occupancy rasters can not be used to predict the layers heights, but are used purely as a complementary supervision to the point labels. 

In Table~\ref{tab:height_mae}, we report the precision of the reconstructed height maps for all layers. The estimation is almost perfect for the top of the overstory, which is expected as the top of this layer can be reconstructed using the highest points. The other values are harder to estimate, particularly the top of the understory layer. Nonetheless, the MAE is overall below $1\:$m ($\sim 25$\% MRE), which constitutes a solid baseline for such a complicated regression task.
Both regression algorithms yield poor performance on height estimation, especially for the lower layers. We interpret this by the fact that the baselines operate pixel by pixel while our network has a large receptive field of several meters. We conclude that global spatial features are key for the structured reconstruction of vegetation layers in dense forests. 

\begin{table}
    \caption{{\bf Height Regression Performance.} We report the mean and standard deviation of the height of the top and bottom of different vegetation layers of our validation set. We compare the Mean Average Error (m) and Mean Error Ratio (\%) of the predictions of our approach and two baseline regression models.}
    \label{tab:height_mae}
    \centering
    \begin{tabular}{clcccc}
        \toprule
                    &&\multirow{1}{*}{\small Gr. veg.}
                    & \multirow{1}{*}{ \small Under.}    & \small \small Over.  & \small Over.
                    \\
                    &&top&top & base & top\\
        \midrule
        \multicolumn{2}{c}{\multirow{2}{*}{Height, m}}  & 0.9   & 2.7 & 11.3 & 17.3 \\
                                          &             & $\pm$0.3 & $\pm$ 1.1 & $\pm$ 5.2 & $\pm$ 4.0\\
        %\multicolumn{2}{c}{Height, m}       & \small{0.9$\pm$0.3} & \small{2.7 $\pm$ 1.1} & \small{11.3 $\pm$ 5.2} & \small{17.3  $\pm$ 4.0}\\
        \midrule
        \multirow{3}{*}{\rotatebox{90}{MAE, m}}&Ours              & \bf 0.03 & \bf 0.3 & \bf 1.1 & \bf 0.1\\
        &Lin. Reg.         & 0.8 & 2.2 & 6.9 & 3.5\\
        &RF Reg.           & 0.3 & 2.0 & 7.1 & 3.3\\\midrule
        \multirow{3}{*}{\rotatebox{90}{\;\;\;\;MRE, \%\!\! }}&Ours             & \bf 2.9 & \bf 22.0 & \bf 26.5 & \bf 0.7\\
        &Lin. Reg.         & 81.6 & 111.7 & 61.5 & 24.3\\
        &RF Reg            & 35.8 & 85.0 & 64.4 & 22.7\\
        \hline
    \end{tabular}

\end{table}

%........................................
\subsection{Ablation Study}
\label{sec:ablation_study}
%........................................
\begin{table*}
    \caption{{\bf Quantitative Ablation Study.}  Quantitative impact of some design and parameter choices.}
    \label{tab:ablation}
    \centering
    \begin{tabular}{llccccccc}
        \hline
        Model& Param.&\multicolumn{4}{c}{MRE, \%}
         &
        ~
        & \multicolumn{2}{c}{mIoU, \%}\\
        \cline{3-6}\cline{8-9}

         &&\multirow{2}{*}{Ground veg.}& \multirow{2}{*}{Understory}  & Overstory & Overstory && \multirow{2}{*}{2D}  & \multirow{2}{*}{3D}\\
          &&               &             & base           &top            &   & \\
        \hline
        % Ours                            & 3.67   & 24.61  & 12.69 & 0.46 & 80.62  & 53.47\\
        Our Parameterization&-                           & 2.9   & 22.0  & 26.5 & 0.7 && 80.0  & 53.9\\%ok
        a) No 2D Supervision&$\lambda=0$                            & -     & 32.6  & 25.1 & 0.7 && 44.4  & 56.7\\%ok
        b) No  Elevation Modeling& $\mu = 0$            & 3.1   & 25.1  & 21.7 & 0.7 && 79.2  & 56.7\\%ok
        b) Higher Elevation Modeling&  $\mu = 0.5$      & 4.0   & 28.8  & 26.4 & 0.7 && 80.6  & 51.7\\%ok
        b) Sparser Point Sampling& $S=8192$                 & 2.6   & 26.6  & 31.3 & 0.7 && 79.3  & 51.0\\%ok
        d) Smaller Cylinders& $R=2\:$m                        & 3.1   & 26.0  & 28.7 & 0.6 && 82.7  & 55.4\\%ok
        d) Larger Cylinders& $R=10\:$m                        & 3.9   & 29.5  & 30.2 & 0.7 && 77.6  & 52.8\\%ok
        
        e) Finer Raster Resolution& $r=0.25\:$m                     & 1.1   & 15.5  & 25.3 & 0.8 && 79.7  & 56.9\\%ok
        e) Coarser Raster Resolution& $r=1\:$m                        & 20.0  & 64.5  & 30.3 & 6.8 && 70.1  & 59.5\\%ok
        e) Coarser Raster Resolution& $r=2\:$m                        & 37.7  & 99.5  & 47.5 & 19.8 && 59.4  & 56.4\\%ok
        % Only $XYZ$                      & Ground veg.   & Understory  & Overstory base & Overstory top & 2D  & 3D\\
        \hline

    \end{tabular}
\end{table*}

\begin{figure*}
    \centering
    \begin{tabular}{ccc|cc}
    \begin{subfigure}{.191\textwidth}
        \includegraphics[width=\linewidth]{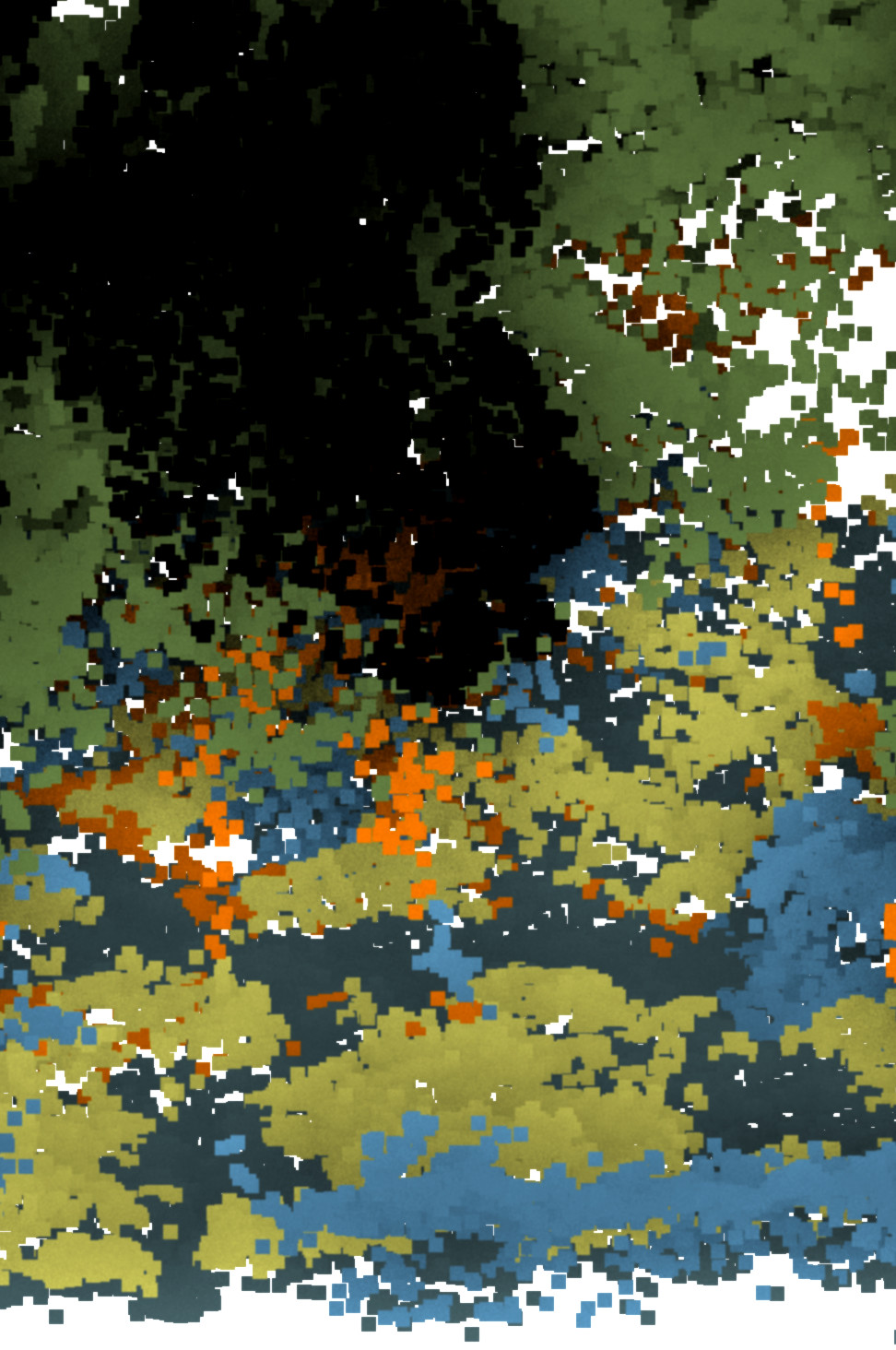}
    \caption{Ours.}
    \label{fig:ablation:a}
    \end{subfigure}
         &
    \begin{subfigure}{.191\textwidth}
    
    \begin{tikzpicture}
        \node[anchor=south west,inner sep=0] (image) at (0,0) { \includegraphics[width=\linewidth]{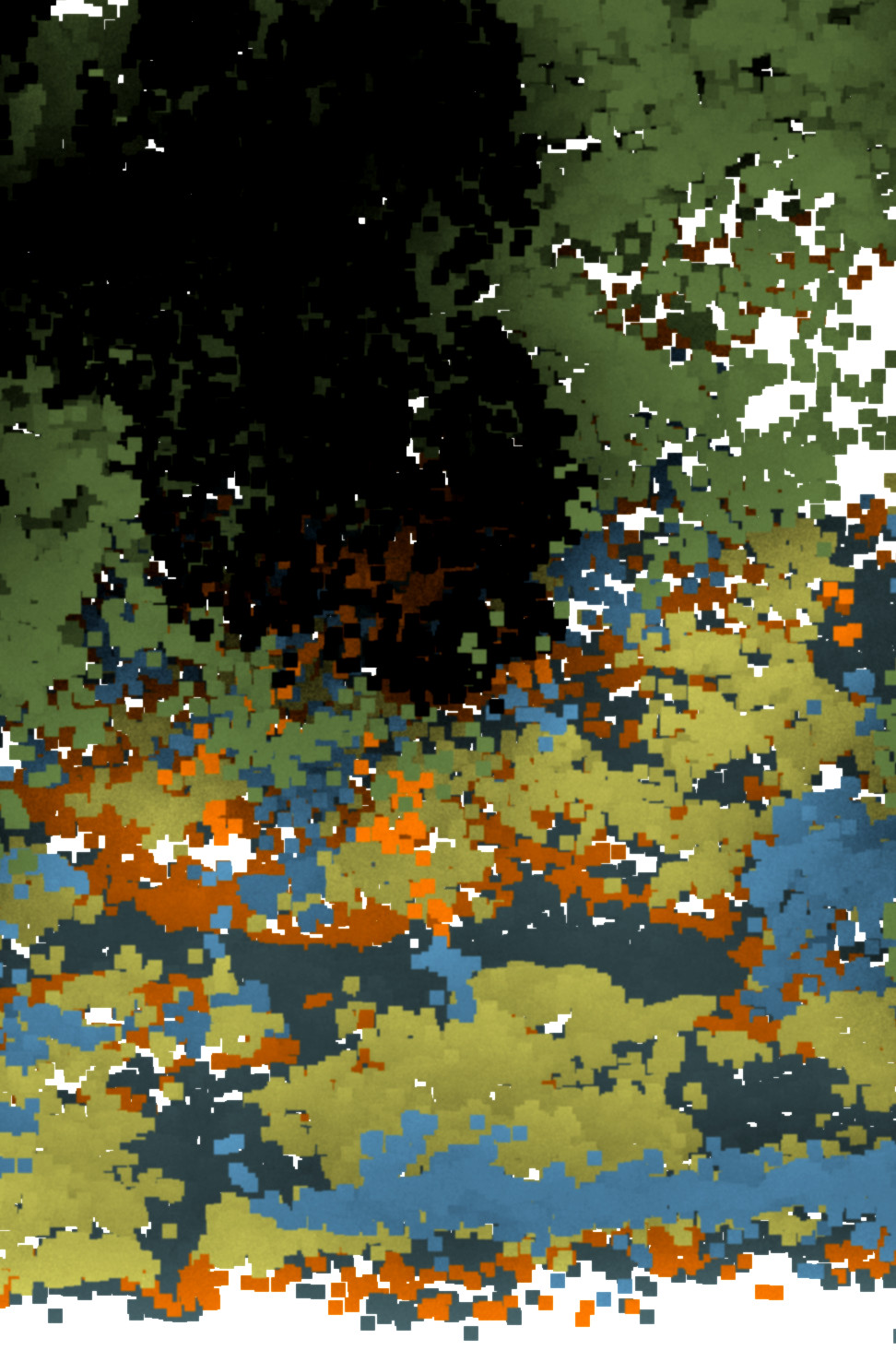}
          };
        \begin{scope}[x={(image.south east)},y={(image.north west)}]
        \draw[draw=red, ultra thick] (0.6,0.07) ellipse (1.3cm and 2mm);
        \draw[draw=red, ultra
        thick, rotate around={55:(0.75,0.35)}] (0.75,0.35) ellipse (4mm and 8mm);
        % \draw[draw=red, ultra
        % thick, rotate around={45:(0.75,0.35)}] (0.75,0.35) ellipse (3mm and 8mm);
        % \draw[draw=red, ultra
        % thick, rotate around={-75:(0.2,0.25)}] (0.2,0.25) ellipse (3mm and 6.5mm);
        % \draw[draw=red, ultra
        % thick, rotate around={-45:(0.2,0.25)}] (0.2,0.25) ellipse (3mm and 6mm);
        \end{scope}
    \end{tikzpicture}
        
    \caption{No elevation modeling.}
    \label{fig:ablation:b}
    \end{subfigure}
         & 
    \begin{subfigure}{.191\textwidth}
    \begin{tikzpicture}
        \node[anchor=south west,inner sep=0] (image) at (0,0) {
        \includegraphics[ width=\linewidth]{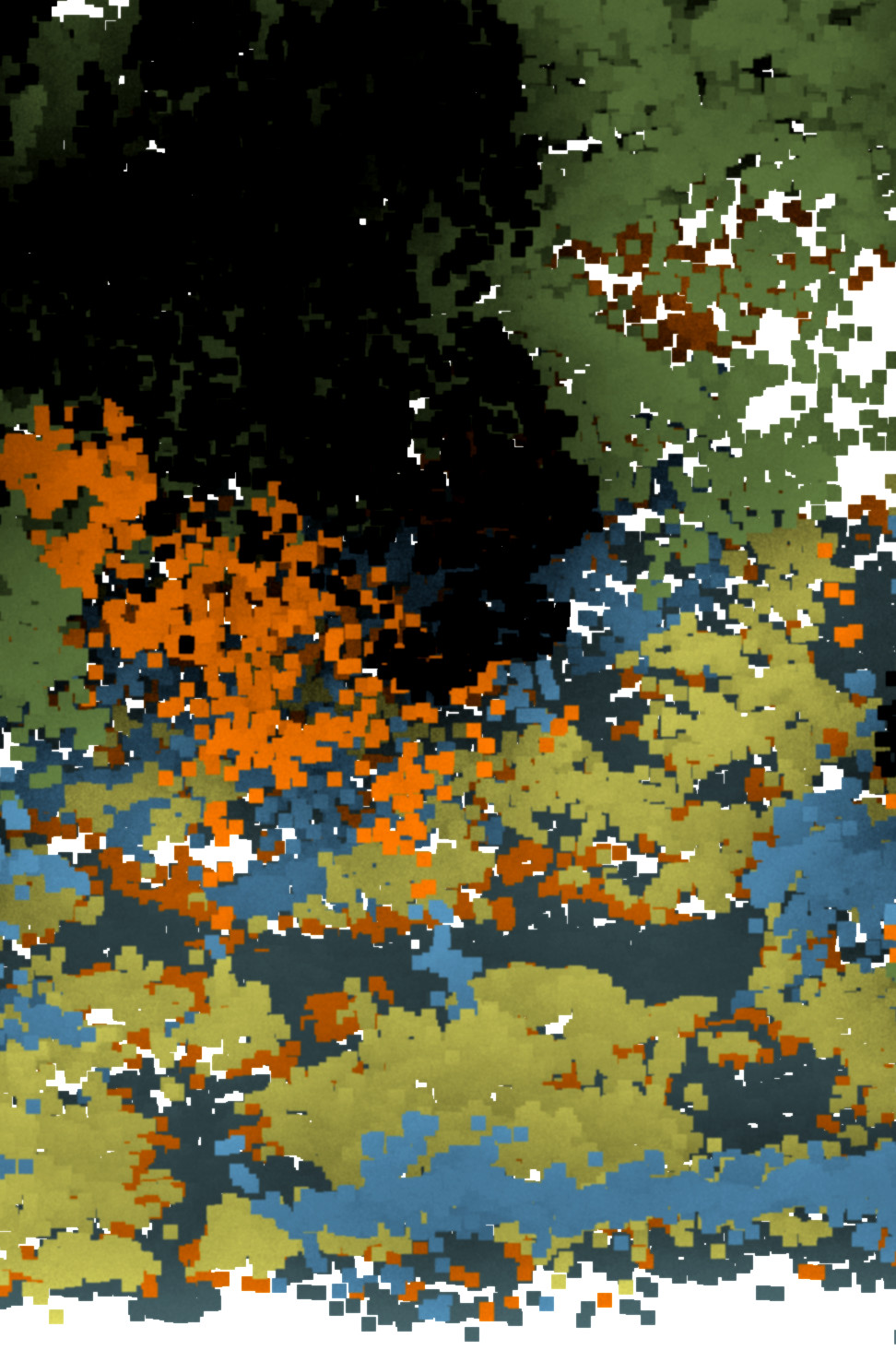}
              };
        \begin{scope}[x={(image.south east)},y={(image.north west)}]
        \draw[draw=red, ultra thick, rotate around={60:(0.75,0.28)}] (0.75,0.28) ellipse (4mm and 6mm);
        \draw[draw=red, ultra thick, rotate around={-20:(0.2,0.25)}] (0.2,0.25) ellipse (4mm and 8mm);
        \end{scope}
        \end{tikzpicture}
    \caption{Sparser sampling.}
    \label{fig:ablation:c}
    \end{subfigure}
    &
    \begin{subfigure}{.126\textwidth}
        \includegraphics[ width=\linewidth]{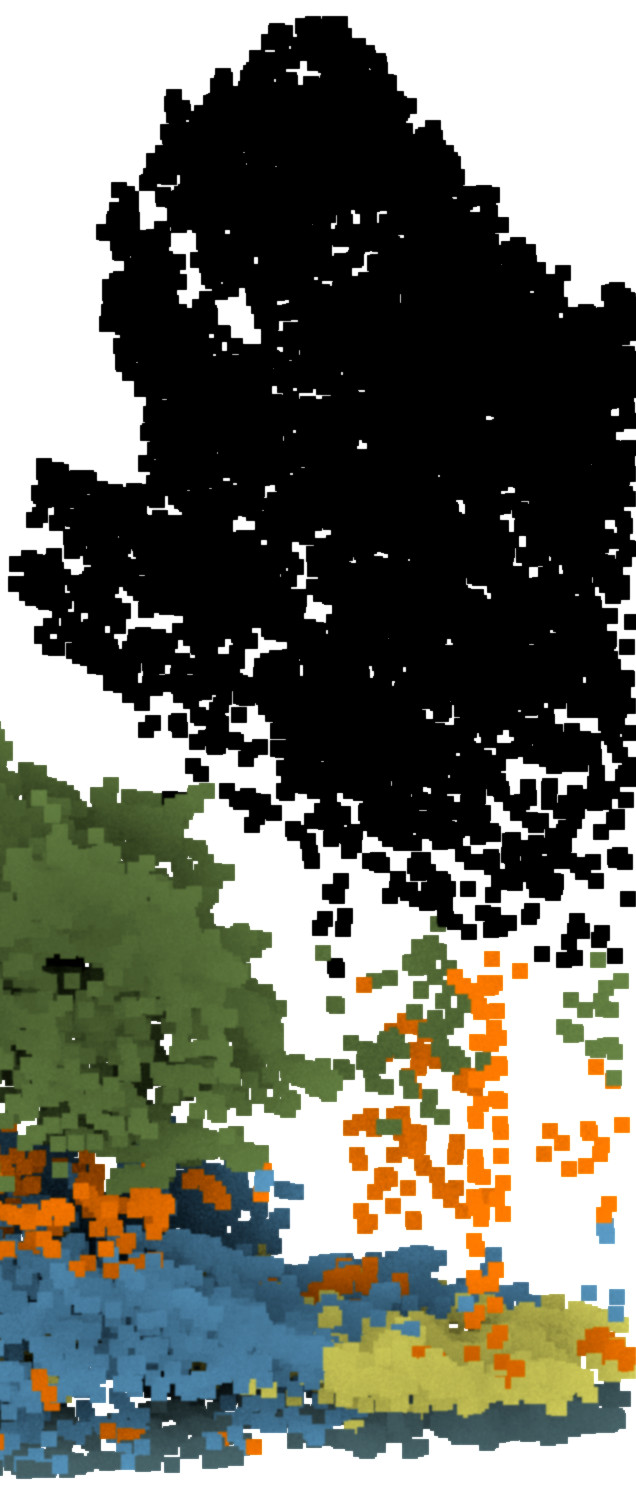}
    \caption{Ours.}
    \label{fig:ablation:d}
     \end{subfigure}
        & 
    \begin{subfigure}{.126\textwidth}
    \begin{tikzpicture}
        \node[anchor=south west,inner sep=0] (image) at (0,0) {
        \includegraphics[ width=\linewidth]{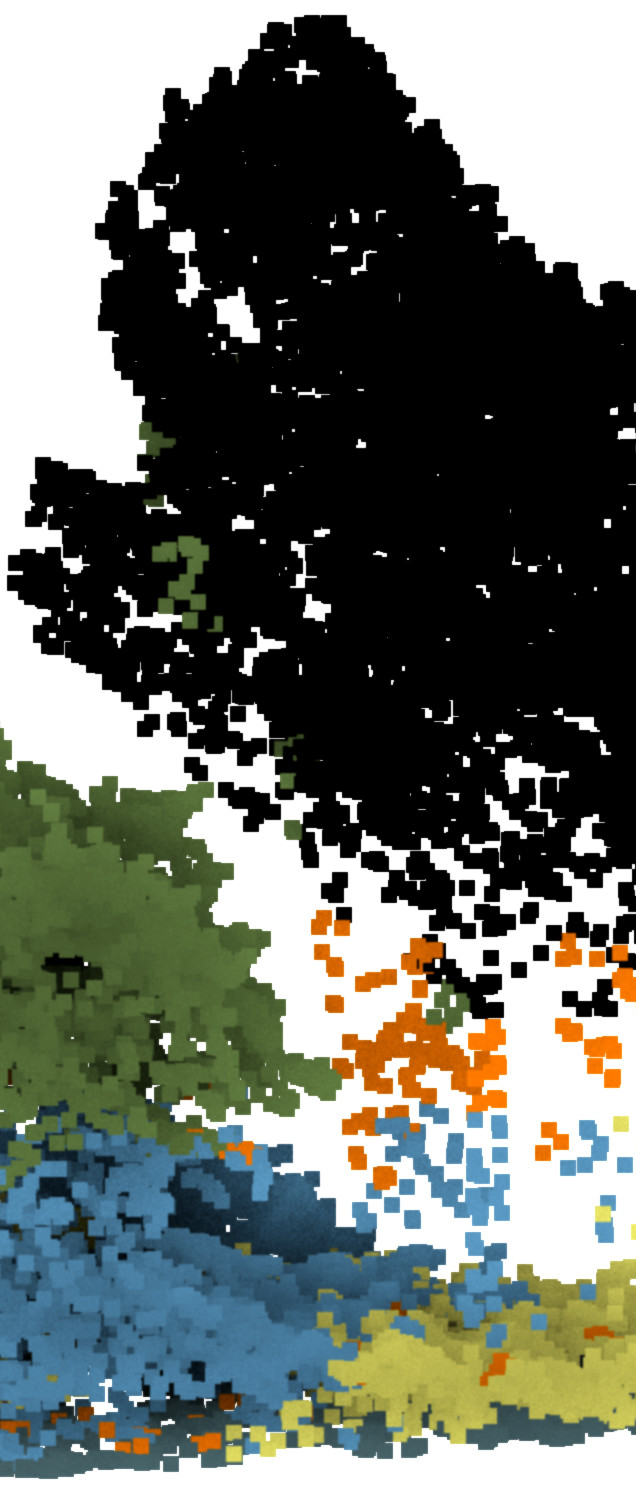}
              };
        \begin{scope}[x={(image.south east)},y={(image.north west)}]
        \draw[draw=indigo, ultra thick] (0.68,0.18) circle (5mm);
        \draw[draw=indigo, ultra thick] (0.2,0.2) ellipse (4mm and 2mm);
        \end{scope}
        \end{tikzpicture}
    \caption{Small Cylinders.}
    \label{fig:ablation:e}
    \end{subfigure}

    \end{tabular}
    % \caption{{\bf Qualitative Ablation Study.} Illustration of the impact of the absence of elevation modelling \subref{fig:ablation:b}} and sparser sampling \subref{fig:ablation:c}, among other, we observe in both results that many shrub points are classified as stems, even when no real tree stem is close, contrary to our model \subref{fig:ablation:a} that has better classification quality of close-to-ground points. }
    \caption{{\bf Qualitative Ablation Study.} Illustration of the impact of the absence of elevation modeling \subref{fig:ablation:b}, sparser sampling \subref{fig:ablation:c} and smaller cylinders \subref{fig:ablation:e}. Among other inaccuracies, we observe in both \subref{fig:ablation:b} and \subref{fig:ablation:c} that many ground vegetation points are erroneously classified as stems
    \protect\tikz \protect\draw [red, thick] (0,0) circle (1mm); . 
    In \subref{fig:ablation:e}, we observe that smaller cylinder size ($R=2$m)
    leads to misclassify tree stems as understory \protect\tikz \protect\draw [indigo, thick] (0,0) circle (1mm); . 
    %In \subref{fig:ablation:e}, we observe the classification results of coniferous trees for the model with smaller cylinder size ($R=2$m) compared to our model~\subref{fig:ablation:e}, those trees are usually easy to identify due to their distinctive shape and absence of understory vegetation underneath. However, we can see that their stems are classified as understory vegetation. Moreover, no tree stems were detected for the coniferous trees (on the left).
    }
    \label{fig:ablation}
\end{figure*}

% \begin{figure}[htp]
% \centering
%     \begin{subfigure}{.32\linewidth}
%         \includegraphics[width=\linewidth]{images/results_pl15_no_el_ours_clip.jpeg}
%     \caption{Our model.}
%     \label{fig:ablation:a}
%     \end{subfigure}
%          \vspace{1pt}
%     \begin{subfigure}{.32\linewidth}
%         \includegraphics[width=\linewidth]{images/results_pl15_no_el_their_clip.jpeg}
%     \caption{No el. modeling.}
%     \label{fig:ablation:b}
%     \end{subfigure}
%          \vspace{1pt}
%     \begin{subfigure}{.32\linewidth}
%         \includegraphics[ width=\linewidth]{images/results_pl15_ss8192_their_clip.jpeg}
%     \caption{Sparser sampling.}
%     \label{fig:ablation:c}
%     \end{subfigure}

% \medskip

%     \begin{subfigure}{.15\textwidth}
%         \centering
%         \includegraphics[ width=0.66\linewidth]{images/results_pl15_R2_ours_clip.jpeg}
%     \caption{Our model.}
%     \label{fig:ablation:d}
%      \end{subfigure}
%         \quad 
%     \begin{subfigure}{.15\textwidth}
%         \centering
%         \includegraphics[ width=0.66\linewidth]{images/results_pl15_R2_their_clip.jpeg}
%     \caption{Smaller Cylinders.}
%     \label{fig:ablation:e}
%     \end{subfigure}

% \caption{{\bf Qualitative Ablation Study.} Illustration of the impact of XXX \subref{fig:ablation:a}}\label{fig:ablation}
% \end{figure}

We now evaluate the impact of our model design choices with quantitative results in Table~\ref{tab:ablation} and qualitative illustrations in Figure~\ref{fig:ablation}. %Since we only have partial annotations, some models have quantitative results similar to ours, however, a visual analysis of the results show the contrary.
We compare our main model with the following configurations.\\
\noindent {\bf a) No 2D Supervision: } we set $\lambda=0$ in Eq~\ref{eq:loss}. Since there is no 3D annotation of the ground vegetation, the prediction for this layer cannot be performed, significantly impacting the prediction of the understory as well. Direct supervision of the occupancy raster proves to be necessary.

\noindent {\bf b) Elevation Modeling: } we remove $\mathcal{L}_{el}$ by setting $\mu=0$ in Eq~\ref{eq:loss}. The performance is not significantly affected. However, a close visual estimation (Figures~\ref{fig:ablation:a}~and~\ref{fig:ablation:b}) show an increasing number of confusions for ground and ground vegetation points, which are erroneously classified as stems.
In contrast, when setting $\mu=0.5$ instead of $0.1$, the height precision of the estimation of both ground vegetation and understory decreases.

\noindent {\bf c) Sparser Point Sampling: } we lower the number of points per cylinder to $S=2^{13}=8192$. Due to the lower point density of close-to-the-ground vegetation, especially under tall trees, sparser point sampling results in errors in the ground vegetation and understory layers, and tree stems are poorly classified, as seen in Figure~\ref{fig:ablation:c}).

\noindent {\bf d) Size of Cylinders: } we set the sampled cylinders radius $R$ to $10\:$m while subsampling the same amount of points. A lower density leads to a similar performance decrease than the previous experiment. However, increasing the subsampling size for $R=10$ would require a significant increase in GPU memory consumption. 

We set $R=2\:$m and adjust the number of subsampled points to to $S=2^{12}=4096$ as well as corresponding parameters in the network.
The resulting cylinders are tall and thin, which do not match well with the spherical feature aggregation scheme of PointNet++, and yield to overall lower performance (Figure~\ref{fig:ablation:e})).
%.
%Hence, we observe (Figure~BLABLA) that many "easily detactable" isolated tree stems are classified as understory as the algorithm can not relate the stem to the tree crown features.\\

\noindent {\bf e) Raster Resolution:} we try different pixel sizes for the occupancy and height rasters: $r=0.25$, $1$, and $2\:$m.
Small pixel sizes give good scores, but the ground truth has a higher proportion of no label pixels. In practice, this results in occupancy and height maps that are less easily visually interpretable. Larger pixels lead to lower precision as their size exceeds the scale of some individual instances.
%produces overdetailed rasters that are more difficult to interpret, the higher generalization level influence the prediction quality of ground vegetation and understory as its footprint can not be correctly represented in the 2D ground truth raster maps.\\

\vspace{-2mm}
\section{Conclusion}

In this paper, we introduced WildForest3D a novel airborne LiDAR dataset of dense forest with fine-grained 3D annotations. We also presented a deep learning algorithm for the structured analysis of vegetation layers from large 3D LiDAR point clouds. Contrary to existing approaches, our algorithm is able to provide high precision vegetation occupancy maps for different layers along with their corresponding height. 
We hope that by releasing our data and code in open source we will encourage the computer vision community to consider the challenging and impactful problem of automated forest data analysis at a large scale. 

In future works we will modify our model to operate on complementary data sources such as high resolution aerial images,  multispectral satellite image time series, and contextual information.
%In perspective, we plan to improve the identification of some classes, such as tree stems. Moreover, we are planning to introduce a multi-source approach by adding very high resolution aerial images. It will increase the number of detected tree classes, as well as allow us do identify individual tree crowns in overstory layer.

%%%%%%%%% REFERENCES
\FloatBarrier
{\small
\bibliographystyle{ieee_fullname}
\bibliography{egbib}
}

\end{document}